%% file: ms.tex
\documentclass[a4paper]{interact}

\usepackage{lineno,hyperref}
\modulolinenumbers[5]
\usepackage{soul}

\usepackage{amsmath,amssymb,amsfonts}
\usepackage{algorithmic}
\usepackage{graphicx}
\usepackage{textcomp}
\usepackage{xcolor}
\def\BibTeX{{\rm B\kern-.05em{\sc i\kern-.025em b}\kern-.08em
    T\kern-.1667em\lower.7ex\hbox{E}\kern-.125emX}}
\usepackage{enumitem}
\usepackage{longtable}
\usepackage{booktabs}
\usepackage{multirow}
\usepackage{dsfont}
\usepackage{adjustbox}
\usepackage{graphicx}
\usepackage{caption}
\usepackage{subcaption}

\usepackage{amsthm}

\usepackage[linesnumbered,ruled]{algorithm2e}
\usepackage{amssymb}
\usepackage{ifthen}
\usepackage{amsmath}
\usepackage{amsfonts}
\usepackage{booktabs}
\newlength{\Oldarrayrulewidth}

\usepackage{color, colortbl}
\definecolor{Gray}{gray}{0.9}

\usepackage{changepage}

\begin{document}

\title {Accident Impact Prediction based on a deep convolutional and recurrent neural network model}

\author{
    Pouyan Sajadi\textsuperscript{a}$^*$\thanks{$^*$Both authors contributed equally to this work. Email: pouyan.sajadi78@student.sharif.edu. \textsuperscript{a}Department of Industrial Engineering, Sharif University of Technology, Tehran, Iran.},
    Mahya Qorbani\textsuperscript{b}$^*$\thanks{Email: mqorbani3@gstech.edu. \textsuperscript{b}School of Industrial and System Engineering, Georgia Institute of Technology, Atlanta, GA, USA.},
    Sobhan Moosavi\textsuperscript{c}\thanks{Email: moosavi.3@osu.edu. \textsuperscript{c}Department of Computer Science and Engineering, Ohio State University, Columbus, Ohio, USA.},
    Erfan Hassannayebi\textsuperscript{a}$^{\ddag}$\thanks{$^{\ddag}$Corresponding author: Email: hassannayebi@sharif.edu. \textsuperscript{a}Department of Industrial Engineering, Sharif University of Technology, Tehran, Iran.}
}

\date{} 
\maketitle
\begin{abstract}
Traffic accidents pose a significant threat to public safety, resulting in numerous fatalities, injuries, and a substantial economic burden each year. The development of predictive models capable of real-time forecasting of post-accident impact using readily available data can play a crucial role in preventing adverse outcomes and enhancing overall safety. However, existing accident predictive models encounter two main challenges: first, reliance on either costly or non-real-time data, and second the absence of a comprehensive metric to measure post-accident impact accurately. To address these limitations, this study proposes a deep neural network model known as the \textit{cascade} model. It leverages readily available real-world data from Los Angeles County to predict post-accident impacts. The model consists of two components: Long Short-Term Memory (LSTM) and Convolutional Neural Network (CNN). The LSTM model captures temporal patterns, while the CNN extracts patterns from the sparse accident dataset. Furthermore, an external traffic congestion dataset is incorporated to derive a new feature called the ``accident impact'' factor, which quantifies the influence of an accident on surrounding traffic flow. Extensive experiments were conducted to demonstrate the effectiveness of the proposed hybrid machine learning method in predicting the post-accident impact compared to state-of-the-art baselines. The results reveal a higher \textit{precision} in predicting minimal impacts (i.e., cases with no reported accidents) and a higher \textit{recall} in predicting more significant impacts (i.e., cases with reported accidents). 
\end{abstract}



\input{tex/introduction}
\input{tex/literature}

\input{tex/dataset}
\input{tex/relabeling}
\input{tex/method}
\input{tex/Experiments_and_Results}
\input{tex/conclusion}

\input{tex/Interest}
\input{tex/Data_Availability}

\bibliographystyle{unsrt} 
\bibliography{references} 


\end{document}

%% file: tex/introduction.tex
\section{Introduction}
\label{sec:intro}
Traffic crashes are a major public safety concern, leading to significant fatalities, injuries, and economic burdens annually. In 2020, the fatality rate on American roads saw a staggering 24\% increase compared to 2019, the largest year-over-year rise since 1924, according to the National Safety Council (NSC). Understanding the causes and impacts of traffic accidents is crucial for evaluating highway safety and addressing negative consequences, including high fatality and injury rates, traffic congestion, carbon emissions, and related incidents. Despite ample existing literature, the development of accurate and efficient accident prediction models remains a challenging and fascinating research area.

Over the past decades, a wide range of predictive and computational methods have been proposed to predict traffic accidents and their impacts \cite{wang2016soft}. These models progressed from linear to nonlinear models and from traditional predictive regression models \cite{zhang2019prediction, austin2002alternative, Oyedepo2010AccidentPM} to today's most common data analytic and machine learning algorithms \cite{wei2007sequential, K-MEANS, sarkar2019application}. A majority of the studies treat the problem as a standard binary classification task \cite{wenqi2017model, ozbayoglu2016real}, with the output being a categorical event (accident or non-accident) and with no information on the post-accident consequences. Furthermore, existing studies that investigated post-accident impacts mostly failed to provide a compelling way to define and measure the ``impact'', which is a significant challenge. 
The commonly used metric to measure and illustrate the impact of accidents on traffic flow is ``duration''. However, to our understanding, this metric is not ideal because it can be influenced by various environmental factors, including the type of road and accessibility conditions. For example, if a road has accessibility issues, it may take longer to clear an accident scene, resulting in a longer duration. Furthermore, the datasets typically utilized for these studies are not publicly available and cannot be employed in real-time projects.

In recent years, hybrid neural networks that combine the strengths of multiple neural network components have garnered considerable attention in academic circles. To address the aforementioned issues and propose a practical solution for predicting the impact of accidents, we suggest the use of a LSTM-CNN cascade model. This approach harnesses the capabilities of recurrent neural networks (RNNs), specifically Long Short-Term Memory (LSTM), and convolutional neural networks (CNNs). LSTM is well-known for its ability to handle temporal relationships and capture long-term dependencies, while CNN excels at extracting spatial and temporal patterns from spatiotemporal event data, such as accident events \cite{li2020real}. Moreover, we introduce a novel metric that defines the impact of accidents on surrounding traffic flow, which addresses an additional challenge that existed in previous studies.

LSTM and CNN models are sequentially combined  in terms of a ``cascade'' model. We specifically chose these two models because of their ability to handle temporal dependencies effectively. Furthermore, their combination offers a stronger capability to distinguish between accident and non-accident events, especially in imbalanced datasets, compared to using each model individually. Our model performs better in discriminating high-impact accidents. It achieves this by first distinguishing between accident and non-accident events and subsequently predicting the post-accident impact of the accident events. In more details, the cascade model utilizes a sequence of the last \(T\) time intervals (e.g., a two-hour interval) within each specific \textit{region} (e.g., a zone specified by a square of size 5km × 5km on map) to predict the probability of an accident and its post-impact in that region for the next time interval (i.e., \(T+1\)). 

This study utilizes four complementary, easy-to-obtain, real-world datasets to build and evaluate the proposed prediction framework. The first dataset provides real-world accident data across the United States; it contains location, time, duration, distance, and severity. The second dataset offers detailed weather information such as weather conditions, wind chill, humidity, temperature, etc. The third dataset contains spatial attributes for each district (e.g., the number of traffic lights, stop signs, and highway junctions). The fourth dataset is a traffic congestion dataset that provides real-world congestion data across the United States. The congestion dataset is also used to develop a novel target feature, {\it gamma}, representing accident impact on its surrounding area. We propose a data-driven solution to define gamma as a function of three accident-impact-related features, \textit{duration}, \textit{distance}, and \textit{severity}. Generally, non-accident intervals are far more prevalent in the real world compared to accident intervals (since accidents are relatively infrequent events). Therefore, working with such an imbalanced dataset poses challenges for model development and evaluation. To address this issue, we employ a random under-sampling method to mitigate the imbalanced nature of the dataset. Using these settings and through extensive experiments and evaluations, we demonstrate the effectiveness of the cascade model in predicting post-accident impact compared to state-of-the-art models that were evaluated using the same datasets. It is worth noting that the datasets used in this research are generally easy to obtain, regarding how they have been collected (more details can be found in \cite{moosavi2019short}) and the replicability of the data collection processes for other regions.

The main contributions of this paper are summarized as follows:
\begin{itemize}
    \item \textbf{A real-world setup for post-accident impact prediction}: This paper proposes an effective setup to use easy-to-obtain, real-world data resources (i.e., datasets on accident events, congestion incidents, weather data, and spatial information) to estimate accident impact on the surrounding area shortly after an accident occurs. Also, this framework employs a data augmentation approach to obtain accurate feature vectors from heterogeneous data sources.  

    \item \textbf{A data-driven label refinement process}: This study illustrates a novel feature to demonstrate post-accident impact on its surrounding traffic flow. This feature combines three factors, namely ``severity'', ``duration'', and ``distance'', to create a compelling feature that we refer to as \textit{gamma} in this work. 
    
    \item \textbf{A cascade model for accident impact prediction}: This paper proposes a cascade model to employ the power of LSTM and CNN to efficiently predict the post-accident impacts in two stages. First, it distinguishes between accident and non-accident events, then it predicts the intensity of impact for accident events. 
\end{itemize}
\sethlcolor{yellow}
In summary, the research contributions, both in terms of application and solution framework, are designed to directly connect to the potential reader, the distinguishing aspects and distinctiveness of the accident impact prediction method.

The remainder of the article is organized as follows. Section~\ref{sec:related} provides an overview of related studies, followed by the description of data in Section~\ref{sec:dataset}. Section~\ref{sec:relabeling} describes our accident labelling approach, and then the cascade model is presented in Section~\ref{sec:method}. Section~\ref{sec:Experiments and Results} illustrates experimentation design and results, and Section~\ref{sec:conclusion} concludes the paper by discussing essential findings and recommendations for future studies. 

%% file: tex/literature.tex
\section{Related Work}
\label{sec:related}

Traffic accident analysis and prediction have been extensively studied over the past few decades. Previous work can be classified into two categories: \textit{Accident Risk Prediction} and \textit{Accident Impact Prediction}. In the following we overview some of the important studies in each category. 

\subsection{Accident Risk Prediction}
These studies focused on predicting the risk of an accident itself, not consequences or any features of it. Two primary goals of these studies are 1) predicting the anticipated number of traffic accidents (i.e., accident rate prediction) \cite{caliendo2007crash,najjar2017combining}, or 2) predicting whether an accident will occur on a particular road section or geographical region or not (i.e., accident detection). In the first category, researchers applied regression models to predict a value as accident rate \cite{ren2018deep,yuan2018hetero,chen2016learning}, while binary classification methods are used for the second category \cite{parsa2020toward,lin2020automated,polat2012automatic}. Some studies focused on temporal dependencies between accidents to predict the probability of accidents for the subsequent intervals \cite{bianchi2017recurrent,tian2015predicting,parsa2019applying}. In contrast, some others (e.g., \cite{yuan2018hetero,wang2021graph}) also sought to consider spatial dependencies.

As one of the latest studies, Kaffash et al. \cite{kaffash2022road} designed an accident risk map using regression neural network tuned with self-organizing map technique which is able to estimate accident risk along 3008 points in a dual carriageway with more than 90\% average accuracy. 
Yuan and Abdel-Aty \cite{yu2013utilizing} used a support vector machine (SVM) for real-time crash risk evaluation. They used a classification and regression tree (CART) model to select the most important explanatory variables, and fed them to the SVM model with a radial-basis kernel function. They achieved 0.81 as their best model’s Area Under the Curve (AUC) value, tested on I-70 freeway crash data from October 2010 to October 2011 provided by Colorado Department of Transportation, and real-time traffic data collected by 30 Remote Traffic Microwave Sensor (RTMS) radars. 
In \cite{lin2015novel}, a novel feature selection algorithm based on a frequent pattern tree model was implemented to identify all the frequent patterns in the accident dataset, then ranked variables by their proposed metric called Relative Object Purity Ratio (ROPR). Finally, two classification models (k-nearest neighbor model and a Bayesian network) were used to predict real-time accident risk.

With the growing diversity and complexity of data, applying deep neural network (DNN) methods usually resulted in more accurate and robust predictions when compared to statistical methods or traditional machine learning models \cite{singh2020deep,theofilatos2019comparing}.
Moosavi et al. \cite{moosavi2019accident} presented a deep neural network model (DAP) using multiple components, including a recurrent, a fully connected, and an embedding component for accident detection on their country-wide dataset. Li et al. \cite{li2020real} used a long short-term memory convolutional neural network (LSTM-CNN) to predict real-time crash risk on arterial roads using features such as traffic flow characteristics, signal timing, and weather condition. Their proposed model outperformed baseline models based on AUC value, sensitivity, and false alarm rate. 
Chen et al. \cite{chen2016learning} used a set of 300K accident records collected from GPS mobility data to predict a parameter \(g\), defined as the summation of accident severity values in square areas of size $500m \times 500m$ during subsequent hour. They applied a stack denoising autoencoder model to extract latent features and then used a logistic regression model to predict the \(g\) parameter. Bao et al. \cite{bao2019spatiotemporal} proposed a spatiotemporal convolutional long short-term memory network (STCL-Net) for predicting citywide short-term crash risk. Their results indicated that spatiotemporal deep neural network approaches perform better than other models to capture the spatiotemporal characteristics based on their multi-source dataset of New York City crashes. 
Ren et al. \cite{ren2018deep} collected extensive traffic accident data from Beijing in 2016 and 2017. They built a deep neural network model based on LSTM for predicting the number of traffic accidents (they called it ``traffic accident risk index'') occurring in a specific zone during the subsequent time interval. 
One limitation of their study is utilizing the traffic accident count for prediction and not using any other related data (such as traffic flow, human mobility, or road characteristics). In another study, Chen et al. \cite{chen2018sdcae} proposed a novel Stack Denoise Convolutional Auto-Encoder algorithm to predict the number of traffic accidents occurring at the city-level. They used two different datasets, one consisting of accident data and another one consisting of traffic flow collected by vehicle license plate recognition (VLPR) sensors. A summary of above studies can be found in Table \ref{tab:accident_risk}. 
\vspace{10pt}
\sethlcolor{green}
\newcolumntype{P}[1]{>{\centering\arraybackslash}m{#1}}
\begin{table}[ht]
\small
\centering
\caption{A summary of previous works on accident risk prediction}

\begin{adjustbox}{width=1\textwidth,center=\textwidth}
\begin{tabular}
{ P{0.1\linewidth}  P{0.3\linewidth} P{0.15\linewidth} P{0.25\linewidth} P{0.3\linewidth} }
\hline
        \rowcolor{Gray} \textbf{Study} & \textbf{Input features} & \textbf{Predicted output} &
        \textbf{Model} & \textbf{Best modeling results} \\

\hline  
        Moosavi et al. \cite{moosavi2019accident} & \rule{0pt}{4ex} 1.Weather conditions
  2.Time of day
 3.Location
  4.	Road features & Accident risk on different road segments & DNN with recurrent, fully connected and embedding components) &F1 score=0.59 for accident class
F1 score=0.89 for non-accident class
\\
\hline
         Li et al. \cite{li2020real} & \rule{0pt}{4ex} 1.Traffic flow characteristics
2.Signal timing
3.Weather condition
 & Real-time crash risk &
 LSTM-CNN
 & False alarm rate=0.132
AUC=0.932

\\
\hline
         Parsa et al. \cite{parsa2020toward} & 1.Traffic
2.Network
3.Demographic
4.Land use
5.Weather feature
 & Occurrence of accidents &
 eXtreme Gradient Boosting (XGBoost)
 & \rule{0pt}{4ex} Accuracy= close to 100\%, Detection rate = varies between 70\% and 83\%,  false alarm rate is less than 0.4\% .

\\

\hline
         Yu and Abdel-Aty \cite{yu2013utilizing} &1.Crash data
2.Real-time traffic data

 & Crash occurrence &
 SVM with RBF kernel
 & \rule{0pt}{4ex} AUC = 0.74 for all crash type, AUC = 0.80 for Multi-vehicle crashes, AUC = 0.75 for 	Single-vehicle crashes (evaluated on 30\% of the whole dataset)

\\
\hline
         Ozbayoglu et al. \cite{ozbayoglu2016real} &1. \rule{0pt}{4ex}	Average velocity
2.Occupancy
3.The capacity difference between time t and t+1
4.Weekday/weekend
5.Rush hour
 & Crash occurrence &
 Nearest neighbor (NN), Regression tree (RT), feedforward neural network (FNN),
 & Accuracy=95.12 for NN
Accuracy=97.59 for RT
Accuracy=99.79 for FNN

\\
\hline
    \end{tabular}
    \end{adjustbox}
    \label{tab:accident_risk}
\end{table}
\sethlcolor{yellow}
\subsection{Accident Impact Prediction}
The other group of studies focused on analyzing and predicting the post-accident impact on the surrounding area. These studies aim to build or directly use features that best indicate the impact of accidents. Most of the existing studies used the duration or severity of the accident to define impact. They have employed various models for accident impact prediction, ranging from regression methods \cite{khattak1995simple,garib1997estimating,peeta2000providing,yu2012methodology} to neural network-based solutions \cite{wang2005vehicle,vlahogianni2013fuzzy,wei2007sequential}. 
ZHANG et al. \cite{zhang2019prediction} used the data of loop detectors and accidents recorded by police or traffic authorities to find traffic accident duration and defined duration of accident plus clearance time as the impact of an accident. They employed two models for impact prediction, multiple linear regression and artificial neural network (ANN). Yu et al. \cite{yu2016comparison} obtained traffic incident duration occurred on a freeway from binary features like good or bad weather, day or night, disabled vehicle, and peak hour. They compared results obtained from two models, ANN and SVM. The ANN model comprehensively provided better results for long-duration incident cases, while SVM performed better for short and medium-duration incidents. \cite{ma2017prioritizing} implemented gradient boosting decision trees to predict freeway incident clearance time based on different explanatory variables. Compared to baseline models (SVM, NN, and random forest), their model resulted in superior performance in terms of interpretation power and prediction accuracy. \cite{yu2017deep} is one of the few accident impact prediction studies that focused on directly forecasting post-accident traffic flow; however, their data was limited to a single route rather than the entire road network of a city or an entire area. 
In one of the latest studies, \cite{lin2020real} applied three machine learning methods (i.e., SVM, NN, random forest) to predict the
duration of different traffic condition states after traffic accidents and
considered the updating effect by adding newly acquired data to the
prediction. The above studies are summarized in Table \ref{tab:accident_impact}.
\newcolumntype{P}[1]{>{\centering\arraybackslash}m{#1}}
\sethlcolor{green}
\begin{table}[htb]
\small
\centering
\caption{A summary of previous works on accident impact prediction}

\begin{adjustbox}{width=1\textwidth,center=\textwidth}
\begin{tabular}
{ P{0.1\linewidth}  P{0.3\linewidth} P{0.2\linewidth} P{0.15\linewidth} P{0.3\linewidth} }
\hline
        \rowcolor{Gray} \textbf{Study} & \textbf{Input features} & \textbf{Predicted output} &
        \textbf{Model} & \textbf{Best modeling results} \\

\hline
        Zhang et al. \cite{zhang2019prediction} & \rule{0pt}{4ex} 1.Temporal 2.Spatial 3.Environmental 4.Traffic 5.Accident details

 & Total duration and clearance time & Multiple linear regression and ANN& MAPE=27.1\% for total duration,
MAPE=49.8\% for clearance time

\\
\hline
        Yu et al. \cite{yu2016comparison} &\rule{0pt}{4ex}  1.Night 2.Casualties 3.Peak hour 4.Bad weather 5.Facility damage 6.Disabled Vehicle 7.Heavy tow truck 8.Lay-by occupied 9.Hazard material involved 10.Rollover vehicle involved
 & Incident duration &
 ANN and SVM
 & MAPE = 19\%

\\
\hline
         Ma et al. \cite{ma2017prioritizing} & \rule{0pt}{4ex} 1.Accident details 2.Temporal 3.Geographical 4.Environmental 5.Traffic 6.Operational
 & Incident clearance time &
 gradient boosting decision trees (GBDT)
 &\rule{0pt}{4ex} MAPE=16\% for clearance time less than 15min and MAPE=33\% for clearance time more than 15min

\\
\hline
         Rose Yu et al. \cite{yu2017deep} & \rule{0pt}{4ex} 1.Accident type 2.Downstream post mile 3.Affected traffic direction 4.Traffic speed

 & The delay
corresponding to the accident 
 &
 Mixture Deep LSTM
 & MAPE = 0.97

\\
\hline
         Yu and Abdel-Aty \cite{yu2013utilizing} &\rule{0pt}{4ex} 1.Crash data
2.Real-time traffic data

 & Crash occurrence &
 SVM
 & AUC = 0.74 for SVM with RBF for all crash type

\\
\hline
         Lin and Li \cite{lin2020real} &1.Accident details 2.Traffic details 3.Environmental 4.Air quality 5.Geographical
 & TAPI (derived from post-accident congestion level and its duration) &
 NN, SVM, RF
 & MAPE = 5.5\%–53.8\%

\\
\hline
         Wang et al. \cite{wang2005vehicle} &1.Vehicle type 2.Location 3.Time of day 4.Report mechanism
 & Vehicle breakdown duration &
 fuzzy logic (FL) and
ANN

 &\rule{0pt}{4ex} RMSE=24 for FL model
RMSE=19.5 for ANN mode version one
RMSE=24.1 for ANN mode version two
\\
\hline
    \end{tabular}
    \end{adjustbox}
    \label{tab:accident_impact}
\end{table}

\vspace{10pt}

To the best of our knowledge, this study is the first that formulates post-accident impact on the surrounding area  
by exploiting a variety of signals such as duration, severity, and road blockage distance. This is an essential step toward providing a more comprehensive definition of impact, which addresses an important shortcoming in previous studies. 
Additionally, as opposed to those studies that only used data for a small region or a limited set of routes, we implement and evaluate the prediction model based on a large area (i.e., Los Angeles County), resulting in more generalizable outcomes. Lastly, this work is based on easy-to-obtain, public datasets which can be acquired from publicly available resources such as real-time traffic data providers\footnote{Examples of providers are Microsoft BingMaps and MapQuest}, historical weather data providers\footnote{Example of provider is Weather Underground, visit \url{https://www.wunderground.com/history}}, and open-street-map (OSM)\footnote{Visit \url{https://www.openstreetmap.org}}; thus other researchers can conveniently replicate our model and results for comparative purposes. More details about the process of collecting and building our datasets can be found in \cite{moosavi2019short}.

%% file: tex/dataset.tex
\section{Dataset}
\label{sec:dataset}
This section describes all datasets used to build our accident impact prediction framework. In addition to describing original datasets, the processes of data cleaning, transformation, and augmentations are also described.
Lastly, we briefly study accident duration distribution and compare it with findings by other researchers. It is worth noting that accident duration refers to the time-span that takes to clear the impact of an accident, and we believe it is highly correlated with accident impact. Hence further studying it would help to derive valuable insights when building the predictive model. 

\subsection{Data Sources}\label{datas}
\label{source}
The data sources and study areas are related to the Los Angeles county, the most accident-prone county in the United states\footnote{See \url{https://www.iihs.org/topics/fatality-statistics/detail/state-by-state} for more details.}. Four datasets are used to deliberate all factors involved in the aftermath of accident: {\it accident} dataset, {\it congestion} dataset, {\it point of interest (poi)} dataset, and {\it weather} dataset. The following sub-sections describe these datasets in detail. 

\subsubsection{Accident Dataset}
\begin{figure}[h!]
    \centering
    \includegraphics[scale=0.5]{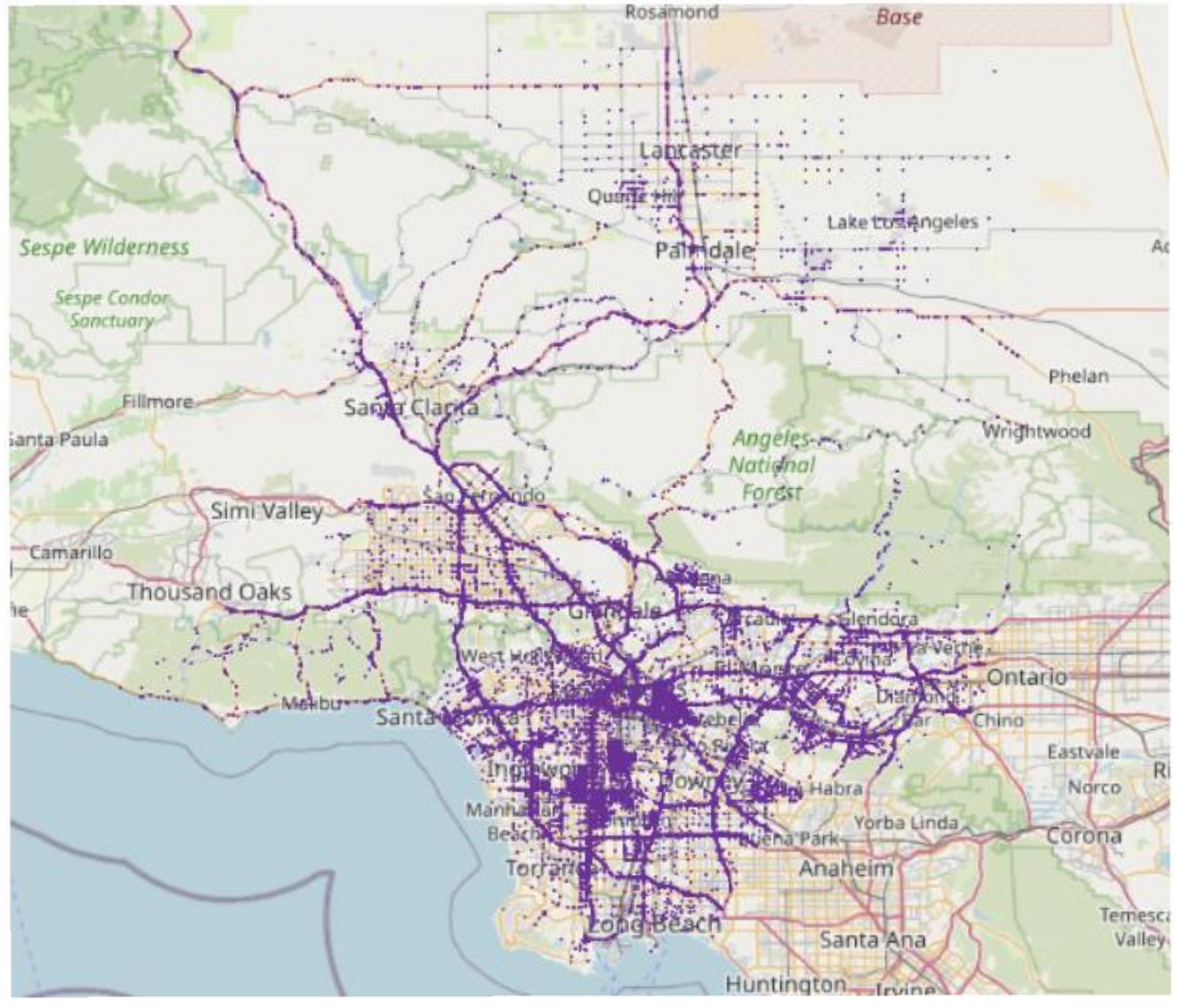}
    \caption{Spatial dispersion of accident locations in Los Angeles County (2016-2020)}
    \label{fig:la_county_accidents}
\end{figure}

This study adopts the US-Accident dataset, a large-scale traffic accident dataset that has been collected from all over the United States \cite{moosavi2019countrywide} between 2016 and 2020. The dataset contains accident events collected from two sources: MapQuest Traffic and Microsoft BingMaps, and it covers 49 states of the United States. 73,553 accident records are extracted that cover four years from August 2016 to December 2020. Figure~\ref{fig:la_county_accidents} shows the dispersion of  accident location in the area of study that covers both freeway and urban arterial roads. Features of reported accidents are listed in Table~\ref{tab:accident_features}. The reader can find a detailed description of features at \url{smoosavi.org/datasets/us_accidents}.

\begin{table}[h!]
    \centering
    \caption{Accident dataset features}
    \begin{adjustbox}{width=1.05\textwidth,center=\textwidth}
    \begin{tabular}{| c | p{0.8\linewidth} |}
        \hline
        \rowcolor{Gray} \textbf{Category} & \textbf{Features} \\
        \hline
        \hline
        Geographical features &  Start latitude, Start longitude, End latitude, End longitude, Distance, Street, Number, Airport code, City, Country, State, Zip code, Side, Amenity, Bump, Crossing, Give way, Junction, No-exit, Railway, Roundabout, Station, Stop, Traffic Calming, Traffic Signal, Turning Loop\\
        \hline
        Temporal features & Start time, End time, Time zone, Sunrise Sunset, Civil Twilight, Nautical Twilight, Astronomical Twilight \\
        \hline
        Other & ID, Severity, TMC \\
        \hline
    \end{tabular}
    \end{adjustbox}
    \label{tab:accident_features}
\end{table}

\subsubsection{Congestion Dataset} \label{Congestion}
Traffic congestion events are essential to predict and also to measure accident impact. The congestion dataset provided in \cite{moosavi2019short} that includes such incidents over the same period and location is considered as the accident data. The dataset includes over 600,000 congestion events, collected between 2016 and 2020, where each event is recorded when traffic speed is slower than typical traffic speed. In addition to the basic details (such as the time and location of a congestion case), the dataset also offers other valuable details such as delay and average speed (the latter is in the form of natural language description of an event reported by a human agent). \emph{``Severe delays of 22 minutes on San Diego Fwy Northbound in LA. Average speed ten mph.''} is an example of the description attribute, that offers details such as the exact ``delay'' and ``speed'' of the congestion. One can easily extract such details from the free-form text via regular expression patterns. In this work, we use congestion data for accident impact labeling that is thoroughly discussed in section~\ref{sec:relabeling}. 

\subsubsection{POI Dataset}
As previous studies have shown, road features have significant impact on determining the possibility of accidents \cite{moosavi2019accident}. As a result, road-network features must be considered when predicting and measuring the impact of accidents. This study utilizes a POI dataset taken from Open Street Map (OSM) that provides geographical features of each location, such as the number of railroads, speed bumps, traffic signals, and pedestrian crossings. 

\subsubsection{Weather Condition Dataset}
Numerous studies had shown the significant impact of weather condition on accident prediction \cite{xing2019hourly,malin2019accident,fountas2020joint}. This research considers the weather data that had been collected from weather stations located in four airports in Los Angeles area (i.e., LAX \footnote{ Los Angeles International Airport} , BUR\footnote{Hollywood Burbank Airport} , VNY\footnote{Van Nuys Airport}, and KWHP\footnote{Whiteman Airport}). The four airports listed above offer hourly weather characteristics for the period of study.  Each record is represented by a vector $e=\langle \textit{Airport, Date, Hour, Temperature, Wind Chill}$, $\textit{Humidity, Pressure, Visibility, Wind Speed, Wind Direction, Precipitation, Weather Conditions}\rangle$, where the weather conditions and weather directions are categorical features (all their possible values are shown in Table~\ref{tab:categorical_weather}).

\begin{table}[h!]
    \centering
    \caption{Details of categorical weather features}
    \begin{adjustbox}{width=1.05\textwidth,center=\textwidth}
    \begin{tabular}{| c | p{0.8\linewidth} |}
        \hline
        \rowcolor{Gray} \textbf{Features} & \textbf{Unique Values} \\
        \hline
        \hline
        Weather Conditions &  'Mostly Cloudy', 'Scattered Clouds', 'Partly Cloudy', 'Clear',
'Light Rain', 'Overcast', 'Heavy Rain', 'Rain', 'Haze', 'Patches of Fog', 'Fog', 'Shallow Fog', 'Thunderstorm', 'Light Drizzle', 'Thunderstorms and Rain', 'Cloudy', 'Fair', 'Mist', 'Mostly Cloudy / Windy', 'Fair / Windy', 'Partly Cloudy / Windy', 'Light Rain with Thunder'
\\
        \hline
        Wind directions & 'E', 'W', 'CALM', 'S', 'N','SE', 'NNE', 'NNW',  'SSE',  'ESE'', 'NE',  'NW', 'WSW', 'ENE', 'SW', 'SSW', 'WNW' \\
        \hline

    \end{tabular}
    \end{adjustbox}
    \label{tab:categorical_weather}
\end{table}

\subsection{Preprocessing and Preparation}\label{Preprocessing}
This section describes the required steps to preprocess different sources of data that are used in this work. Further, the proposal to build input data by combining different sources, to be used for modeling is decribed. In terms of preprocessing, we can elaborate the following steps:

\begin{itemize}
	\item \textbf{Remove duplicated records}: Since the accident data is collected from two potentially overlapping sources, some accidents might be reported twice. Therefore, we first process the input accident data to remove duplicated cases.  	
	\item \textbf{Fill missing values using KNN\footnote{Nearest Neighborhoods} imputation}: The weather data suffers from missing values for some of features (e.g., Sunrise\_Sunset). Based on two nearest neighbors, we impute missing values with mean (for numerical features) or mode (for categorical features). We use time and location to determine distance when finding neighbors. 
    \item \textbf{Treat outliers}: For a numerical feature $f$, if $f_i > \mu_f+3\sigma_f$ or $f_i <\mu_f-3\sigma_f$, then we replace it with \(\mu_f+3\sigma_f\) or \(\mu_f-3\sigma_f\), respectively. Here \(\mu_f\) and \(\sigma_f\) are mean and standard deviation of feature \(f\), respectively, calculated on all weather records. 
	\item \textbf{Omit redundant features}: We remove those features that satisfy either of the following conditions: 1) correlated features based on Pearson correlation; 2) categorical features with more than 90\% data frequency on a specific value.
    \item \textbf{Discretize data}: After data cleaning, both accident and congestion data are first discretized in space and time. The temporal resolution is 2-hour intervals and spatial resolution is set to 5km×5km squares in uniform grids.
\end{itemize}

A vector like \(l(s.t)\in R^m\) is used to represent the input data at time \(t\) in region \(s\), applied for accident impact prediction\footnote{This vector represents input for the predictive task, and later we describe labeling.}. In this way, we convert raw data to \(l(s.t)\in R^{26} \) which consists of 5 categories of features that include 26 different features obtained from raw datasets. Table~\ref{tab:Selected_Attributes} describes details of features used to create an input vector. In addition, due to the rarity of accident events and data sparsity, we drop those zones for which the total recorded accidents is less than a certain threshold \(\alpha\) during the two years. By setting \(\alpha = 75\), which means dropping zones with less than 75 reported accidents in 2 years (i.e., the accidents which were reported in less than 0.8\% of time intervals), the ratio of time intervals with at-least one reported accident (also known as accident-intervals) to all intervals increases by 4.8\%. In section~\ref{undersampling} an under-sampling method is discussed to balance accident to non-accident intervals ratio to a larger extent. 

\begin{table}[h!]
    \centering
    \caption{Selected features after preprocessing and preparation}
    \begin{adjustbox}{width=1.05\textwidth,center=\textwidth}
    \begin{tabular}{| c | p{0.8\linewidth} |}
        \hline
        \rowcolor{Gray} \textbf{Feature Category} & \textbf{Feature} \\
        \hline
        \hline
        Temporal &  Day of Week, Part  of Day (day/night), Sunrise/Sunset\\
        \hline
        Weather & Weather Condition \\
        \hline
         Accident & Severity, Accident Count, Duration, Distance \\
        \hline
         Congestion & Congestion Count, Congestion Delay \\
        \hline
         Spatial & Geohash Code, latitude, longitude, Amenity, Bump, Crossing, Give way, Junction, No-exit, Railway, Roundabout, Station, Stop, Traffic Calming, Traffic Signal, Turning Loop \\
        \hline
    \end{tabular}
    \end{adjustbox}
    \label{tab:Selected_Attributes}
\end{table}

\subsection{Data Augmentation}
\begin{figure}[h!]
    \centering
    \includegraphics[scale=0.73]{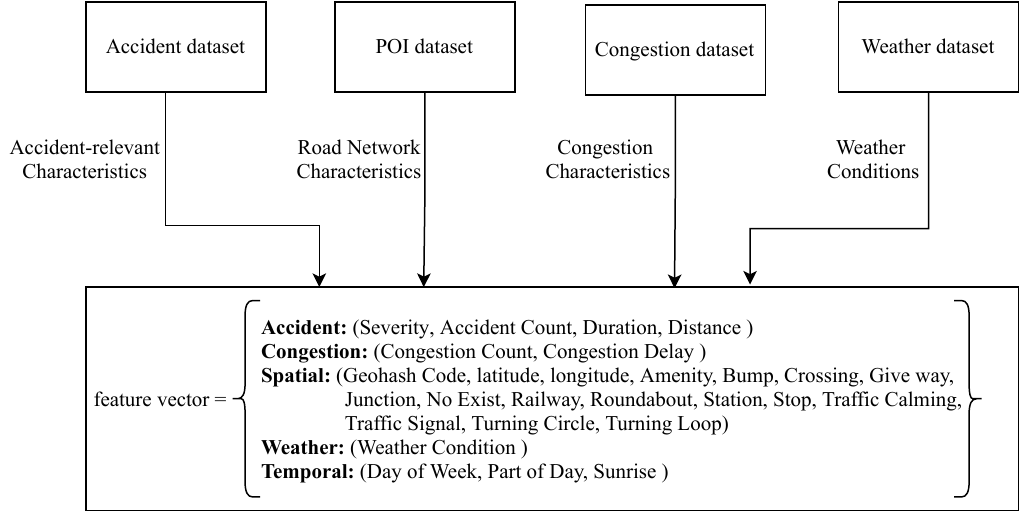}
    \caption{Forming a feature vector at time $t$ in region $s$ based on heterogeneous data obtained from various data sources}
    \label{fig:input feature vector}
\end{figure}

The initial accident dataset includes a variety of features. However, we only use highly accident-relevant features, and further augment the data with additional features described in section~\ref{source} to create the input feature vector. Road-network characteristics of regions are added to differentiate between various types of urban and suburban regions, congestion-related information is added to empower the model in finding latent patterns between accident and congestion events, and weather data collected from the nearest airport based on accident’s occurrence time and location is used to further help the model by encoding weather condition data. Figure~\ref{fig:input feature vector} summarizes how the input feature vector is built in this study. 

To define a proper accident prediction task over geographical areas during different time intervals, it is required to build input data for accident-free time intervals. 
The data are discretized in space and time in order to build these non-accident (or accident-free) intervals (described in Section~\ref{Preprocessing}). When no accident is reported for an interval, accident-related features (i.e., severity, accident count, duration and distance) are set to zero, and remaining features (i.e., geographical, temporal, surrounding congestion and weather condition) are filled accordingly using available resources. 

\subsection{Accident Duration Distribution}
Accident duration is an essential factor to determine impact. Thus, it is worth studying this factor to see how its distribution in our dataset is aligned with datasets that were used in the literature. Previous studies found that a log-normal distribution can best model accident duration  \cite{zhang2019prediction,doane1976aesthetic,skabardonis2003measuring}. 

This section investigates this phenomena using the input data and compares the research findings with \cite{zhang2019prediction}. First, we divide the duration dataset into $K$ classes based on Doane's formula \cite{doane1976aesthetic}: 
$$K=log_{2}n+log_{2}(1+\frac{|g1|}{\sigma_{g1}}) $$
Where \(n\) stands for the total number of observations and \(g1\) refers to the estimated 3rd-moment-skewness of the observations. 
$$\sigma_{g_1}=\sqrt{\frac{6(n-2)}{(n+1)(n+3)}}$$
Class $C_i ,i\in 1,2,\cdots,K$ is specified by its upper bound $(min(D)+i\times\frac{max(D)-min(D)}{K})$ and lower bound $(min(D) + (i-1)\times\frac{max(D)-min(D)}{K})$ where $D$ is accident durations. Then, the frequency of each class is calculated and summation of the squared estimate of errors between frequency of each class and probability density function of fitted distribution is calculated.

\begin{figure}[h!]
    \centering
    \includegraphics[scale=0.8]{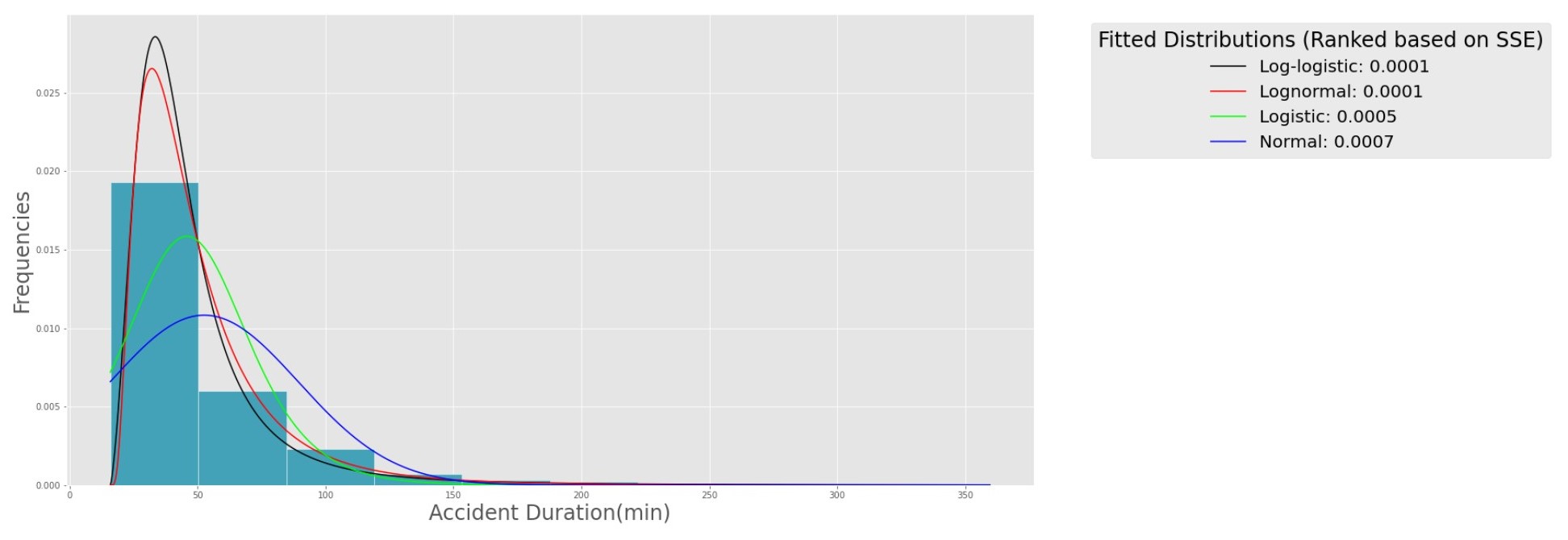}
    \caption{Four best fitted distributions on accident duration data and their summation of squared estimate of errors}
    \label{fig:fitted_distributions}
\end{figure}

The sum of the squared estimate of errors (SSE) calculated for four selected distributions is shown in Figure~\ref{fig:fitted_distributions}. Log-normal and log-logistic distributions have the lowest SSE and therefore describe accident duration data the best. This is aligned with findings reported by Zhang et al. \cite{zhang2019prediction}. They used AIC (Akaike Information Criterion) and the BIC (Bayesian Information Criterion). They found log-normal and log-logistic distributions to be the first and the second best-fitted distributions, respectively.

%% file: tex/relabeling.tex
\section{Label Development of Post-accident Impact}
\label{sec:relabeling}
The previous section discussed data sources, pre-processing, and augmentation steps. However, we have not yet described ``accident impact'' in terms of a quantitative feature or attribute. To the best of our knowledge, there is no exact parameter to show the impact of the accident on its surrounding traffic flow in the real world. There are a number of studies addressing the effect of traffic congestion on accidents \cite{lv2009real,wang2013spatio,sanchez2021understanding}, while the effect of traffic accidents on traffic congestion has been rarely studied \cite{zheng2020determinants,wang2021graph}. Of those few studies, some researches used accident duration as an indicator for accident’s impact, while some others studied the effects of speed changes in surrounding traffic flow as an indicator. We believe none of these approaches are comprehensive and genuine enough to describe how traffic flow would be affected by the occurrence of an accident. 
 
In this study, we propose a novel ``accident impact'' feature based on the delay caused by accidents. This is done by finding a function $\mathcal{F}$ from congestion dataset to estimate delay on accident dataset. In the following , detailed explanation of the process can be found.

\subsection{Accident Impact: A Derived Factor}
There are three attributes in our dataset that can be used to determine ``impact'': 
\begin{itemize}
    \item \textbf{Severity}: this is a categorical attribute that shows severity in terms of delay in free-flow traffic due to accidents. Although it seems to be a highly relevant feature, it suffers from skewness when looking at its distribution (see Figure~\ref{fig:severity}), and being a coarse-grained factor (i.e., it is represented by just a few categorical values). 
    
    \begin{figure}[h!]
        \centering
        \includegraphics[scale=0.4]{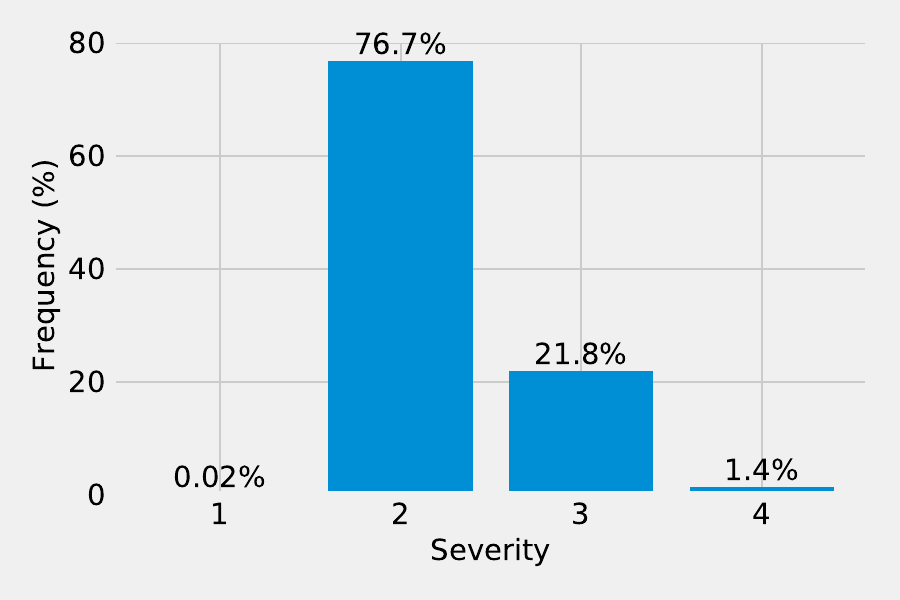}
        \caption{Distribution of Severity values in accident dataset}
        \label{fig:severity}
    \end{figure}

    \item \textbf{Duration}: duration of a traffic accident shows the period from when the accident was first reported until its impact had been cleared from the road network. In this sense, the duration can be considered as another factor to determine impact. Please note that a long duration does not necessarily indicate a significant impact, as it could be related to the type of location and accessibility concerns. But generally speaking, duration is positively correlated with impact. 
    \item \textbf{Distance}: distance shows the length of road extent affected by accident. Similar to duration, a long distance does positively correlate with higher impact. But a high impact accident may not necessarily result in impacting a long road extent.  
\end{itemize}

All three attributes can, to some extent, represent the impact of an accident. However, each alone may not suffice to define the impact primarily. Thus, we propose to build a model that can estimate accident impact given these features as input. 

\subsection{Delay as a Proxy for Impact}
To estimate accident impact, a function $\mathcal{F}$ is defined that maps the three input features to a target value $\gamma$ which we refer it as ``impact'': 
$$\gamma = \mathcal{F}(\textit{Severity}, \textit{Duration}, \textit{Distance})$$
Now the question is: how to find $\mathcal{F}$? While there is no straight-forward approach to estimate $\mathcal{F}$ based on accident data, if we choose to fit it on congestion data (see Section~\ref{Congestion}), we can use some extra signals from the input to estimate the desired function. For congestion events, our data offers a human-reported description of incidents that includes ``delay'' (in minutes) with respect to typical traffic flow (see Table~\ref{tab:cong_description} for some examples). To the best of our knowledge, the ``description'' is generated by traffic officials in a systematic way, therefore it is reliable and accurate. Suppose that the impact in congestion domain is represented by delay (which is a fair assumption given our definition of impact and what delay represents), we can fit $\mathcal{F}$ on congestion data using \textit{delay} as target. Please note that ``delay'' is only available in the congestion dataset, and here the goal is to derive it for accidents and use it as target value.

\begin{table}[h!]
    \centering
    \small
    \caption{Examples of congestion events and features to build function F}
    \begin{adjustbox}{width=1.1\textwidth,center=\textwidth}
    \begin{tabular}{| p{0.06\linewidth} | p{0.6\linewidth} |p{0.09\linewidth} |p{0.1\linewidth} |p{0.1\linewidth} |}
        \hline
        \rowcolor{Gray} \textbf{Event ID} & \textbf{Description} & \textbf{Severity}  & \textbf{Duration (Minute)}  & \textbf{Distance (Mile)} \\
        \hline
        \hline
       1 & \emph{Delays increasing and delays of nine minutes on Colorado Blvd Westbound in LA. Average speed five mph.} & Slow & 49.6 & 2.42\\
        \hline
       2 & \emph {Delays of three minutes on Harbor Fwy Northbound between I-10 and US-101. Average speed 20 mph.} & Moderate & 43.5 & 3.18\\
        \hline 
       3 & \emph{Delays of eight minutes on Verdugo Rd Southbound between Verdugo Rd and Shasta Cir. Average speed five mph.} & Slow & 41.6 & 0.55\\
        \hline 
       4 & \emph{Delays of two minutes on I-5 I-10 Northbound between Exits 132 132A Calzona St and Exit 135A 4th St. Average speed 20 mph.} & Fast & 41.6 & 2.26\\
        \hline         
    \end{tabular}
    \end{adjustbox}
    \label{tab:cong_description}
\end{table}

After fitting $\mathcal{F}$ on congestion events, it is used to estimate impact (represented by $\gamma$) for accidents. It is worth noting that, according to our data, congestion and accident events share the same nature, given their attributes and sources that have been used to collect them. Thus, fitting a function like $\mathcal{F}$ on one and applying it on another is a reasonable design choice. 

\subsection{Estimating Delay}
The reported delays of congestion events are extracted from their description and used as our $\gamma$ variables (i.e., target values). The aim is to find $\mathcal{F}$ by fitting different functions on data and select the best one with lower Mean Squared Error (MSE) and Mean Absolute Error (MAE) values. The overall process is described in Figure~\ref{fig:fittinggamma}. Two models are used to estimate $\mathcal{F}$: 

\begin{figure}[h!]
    \centering
    \includegraphics[scale=0.6]{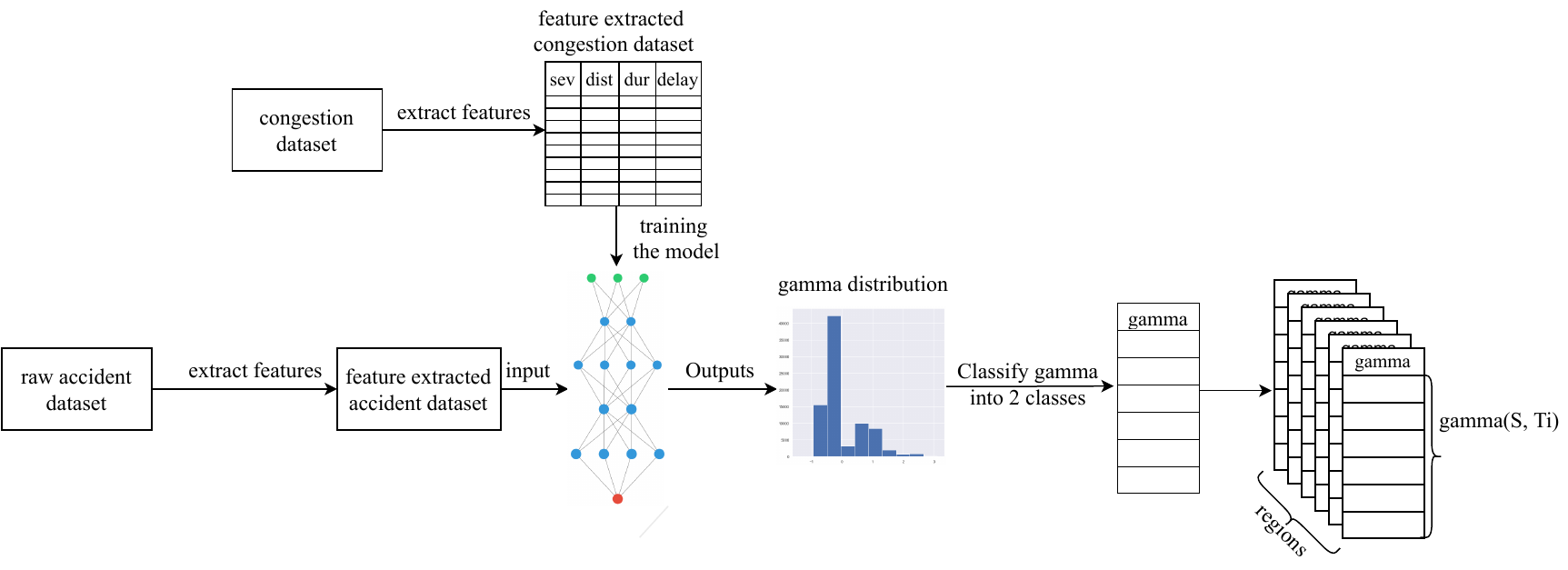}
    \caption{The process of fitting $\mathcal{F}$ on congestion events and applying it on accidents events to predict impact (i.e., $\gamma$)}
    \label{fig:fittinggamma}
\end{figure}

\begin{itemize}
    \item \textbf{Linear regression (LR) model}: given our input features, a linear model seems a natural choice to estimate $\gamma$. So we use a linear regression model for this purpose. 
    \item \textbf{Artificial Neural Network (ANN)}: to examine the impact of non-linearity to estimate $\gamma$, we also used a multi-layer perceptron network with four layers, each with three neurons, and a single neuron in the last layer. We used Adam optimizer \cite{kingma2014adam} with a learning rate of $0.0008$ and trained the model for 200 epochs.
\end{itemize}

To train these models, in total 48,109 congestion records collected between June 2018 and August 2019 are used. The algorithm considers $85\%$ of the data for the training and $15\%$ for evaluation. Table~\ref{tab:models} shows the results of different models to predict delay using the three input features. Based on these results, the non-linear model (i.e., ANN) is selected to estimate the delay (or $\gamma$).  

\begin{table}[h!]
    \centering
    \caption{Result of using ANN and LR models to predict delay (as a proxy for impact) on congestion data}
    \begin{tabular}{| c | c |c |}
        \hline
        \rowcolor{Gray} \textbf{Model} & \textbf{ANN} & \textbf{LR} \\
        \hline
        \hline
       \cellcolor{Gray} \textbf{MSE} & \textbf{0.141} & 0.153\\
        \hline
        \cellcolor{Gray} \textbf{MAE} & \textbf{0.239} & 0.255\\
        \hline
    \end{tabular}
    \label{tab:models}
\end{table}

In what follows, we discuss why it is reasonable to use the function fitted on congestion events and apply it to accident events. First, in our data, there is a resemblance between accidents and congestion events, given their attributes and sources used to collect them. Second, we define the reported delay of congestion as gamma, so with similar inputs from another type of traffic event, the output would be of the nature of delay, which in our opinion is a better indicator of the impact of a traffic accident in comparison to duration or speed changes.

After obtaining $\gamma$ values for accident events, we categorize them into two different classes to show impact of accidents by more detectable labels: ``medium severity'' and ``high severity''; a value of $\gamma$ lower than the median is labeled as ``medium severity'' and a $\gamma$ higher than median is labeled as ``high severity''. Although one could pick a finer-grained categorization of $\gamma$ values (e.g., three or four classes), through the empirical studies, we found the choice of two categories best represent our data.

Please note that $\gamma$ as a real value is not a proper target feature for accident impact prediction since there are numerous contributing factors determining the exact value of accident delay (in seconds or minutes) which cannot be collected in advance. Therefore, the aim is to classify $\gamma$ into two classes (i.e., ``medium severity'' and ``high severity'' ) to deal with this natural drawback of accident impact prediction task.     

%% file: tex/method.tex
\section{Accident impact prediction methodology}
\label{sec:method}
This section describes the proposed model to predict accident impact. The algorithm design is inspired by spatiotemporal characteristics of the input. We leverage two major neural network components: convolutional neural network (CNN) and Long short-term memory (LSTM). While by the former we seek to efficiently encode all types of input attributes, especially the spatial ones, the latter component can efficiently encode temporal aspects of our data and leverage past observations to predict future accident impact. In the remainder of this section, we first briefly introduce some basic concepts, and then discuss the details of the proposed prediction model. 

\subsection{Basic concepts}
\subsubsection{Long short-term memory}
\begin{figure}[h!]
    \centering
    \includegraphics[scale=0.6]{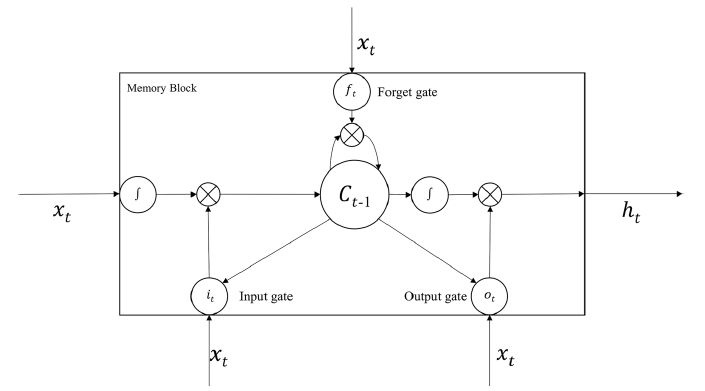}
    \caption{LSTM unit structure}
    \label{fig:LSTMunit}
\end{figure}

The LSTM model proposed by Hochreiter and Schmidhuber \cite{hochreiter1997long} is a variant of the recurrent neural network (RNN) model. It builds a specialized memory storage unit during training through a time backpropagation algorithm. It is designed to avoid the vanishing gradient issue in the original RNN. The key to LSTMs is the cell state, which allows information to flow along with the network. LSTM can remove or add information to the cell state, carefully regulated by structures called gates, including input gate, forget gate and output gate. The structure of a LSTM unit at each time step is shown in Figure~\ref{fig:LSTMunit}. The LSTM generates a mapping from an input sequence vector \(X=(X_1,X_t,\dots,X_N ) \) to an output probability vector by calculating the network units' activation using the following equations ($t$ shows iteration index):
\[i_t=\sigma_g (W_{ix} X_t+W_{ih} h_{(t-1)}+W_{ic} c_{(t-1)}+b_i )\]
\[f_t=\sigma_g (W_{fx} X_t+W_{hf} h_{(t-1)}+W_{cf} c_{(t-1)}+b_f ) \]
\[o_t=\sigma_g (W_{ox} X_t+W_{oh} h_{(t-1)}+W_{oc} c_t+b_o ) \]
\[ c_t=f_t\odot c_{(t-1)}+i_t\odot\sigma_c (W_{cx} X_t+W_{ch} h_{(t-1)}+b_c )\]
\[ h_t=o_t\odot\sigma_g (c_t )\]
\[ y_t=W_{yh} h_{(t-1)}+b_y\]
where $X$ is the input vector, \(W\) and \(b\) are weight matrices and bias vector parameters, respectively, needed to be learned during training. \(\sigma_c\),\(\sigma_g\) are  sigmoid and hyperbolic tangent function, respectively, and \(\odot\) indicates the element-wise product of the vectors. The forget gate \(f_t\) controls the extent to which the previous step memory cell should be forgotten, the input gate \(i_t\) determines how much update each unit, and the output gate \(o_t\) controls the exposure of the internal memory state. The model can learn how to represent information over several time steps as the values of gating variables vary for each time step.
\subsubsection{Convolution Neural Network}
\begin{figure}[h!]
    \centering
    \includegraphics[scale=0.6]{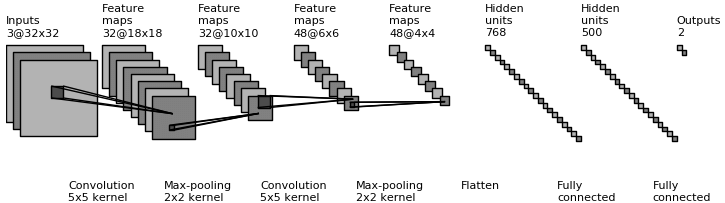}
    \caption{Overall structure of the convolution neural network}
    \label{fig:CNNstructure}
\end{figure}

\begin{figure}
\centering
\begin{subfigure}{.5\textwidth}
  \centering
  \includegraphics[width=0.8\linewidth]{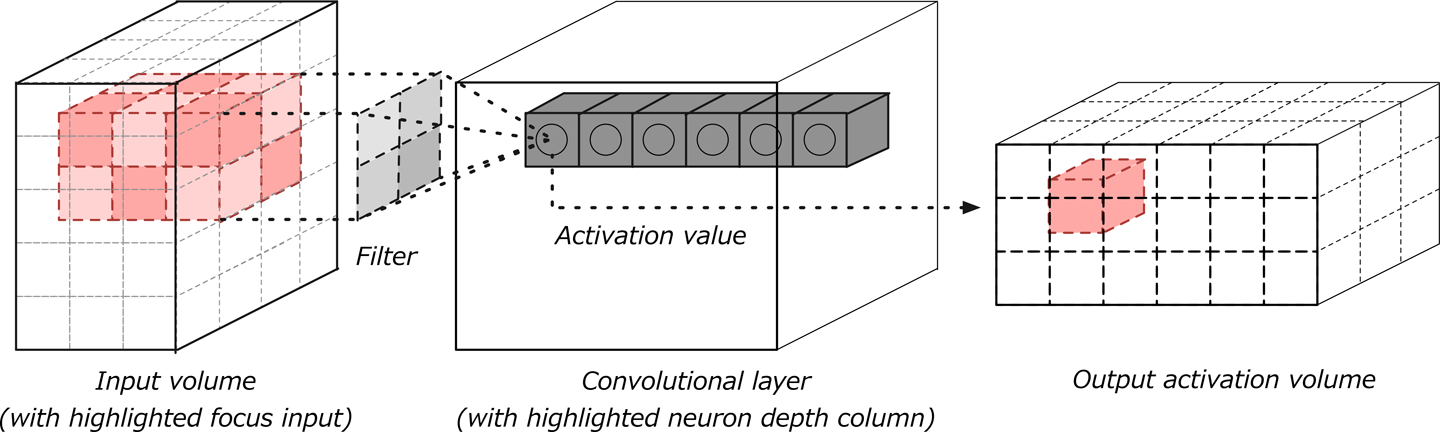}
  \caption{Structure of a convolution layer with input and output volumes} 
  \label{fig:sub11}
\end{subfigure}%
\begin{subfigure}{.5\textwidth}
  \centering
  \includegraphics[width=.6\linewidth]{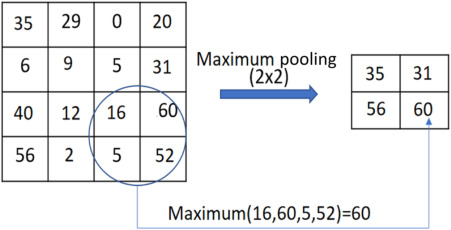}
  \caption{Pooling layer operation} 
  \label{fig:sub12}
\end{subfigure}
\caption{Two main operations in CNN models to extract spatial latent patterns}
\label{fig:CNN_unit2}
\end{figure}

Convolution neural network (CNN) is a multi-layer neural network, which includes two important components for feature extraction: \textit{convolution} and \textit{pooling} layers. Figure~\ref{fig:CNNstructure} shows the overall structure of the convolution neural network. Its basic structure includes two special neuronal layers. The first one is convolution layer; the input of each neuron in this layer is locally connected to the previous layer, and this layer is to extract local features. The second layers is the pooling layer used to find the local sensitivity and perform secondary feature extraction \cite{wang2018prediction}. The number of convolutions and pooled layers depends on the specific problem definition and objectives. Firstly, the model uses a convolutional layer to generate latent features based on the input (Figure~\ref{fig:sub11}). Then, a sub-sampling layer is used on top of the convolutional output to extract more high-level features for the classification or recognition task (Figure~\ref{fig:sub12}). 

\subsection{Model input and output}\label{undersampling}
\begin{figure}[ht!]
    \centering
    \includegraphics[scale=0.55]{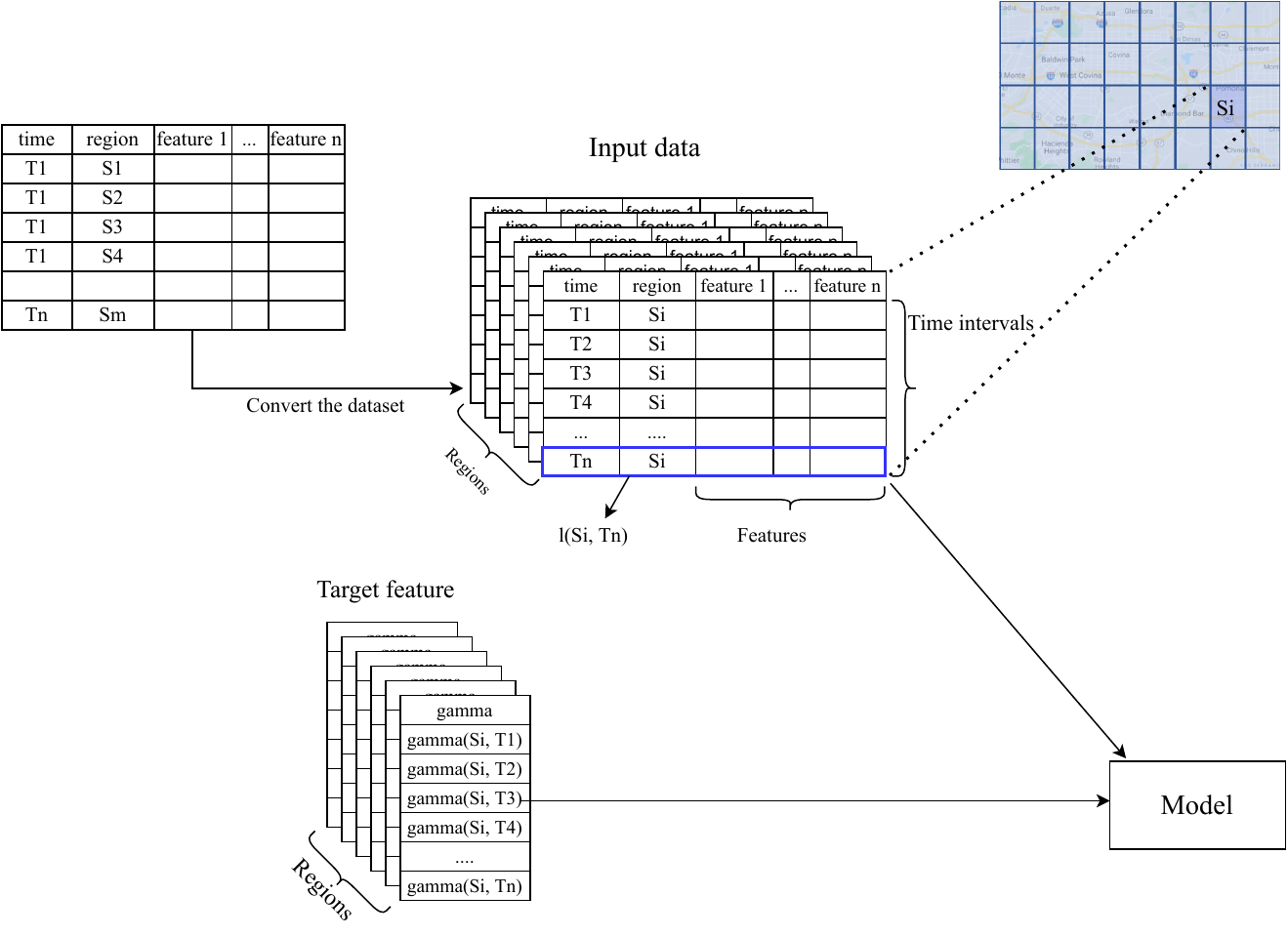}
    \caption{Converting dataset into a 3-D structure of \(l(s_i,t)\) and \(Gammaclass\) in time and space domains. Model inputs and outputs are extracted from this 3-D structure}
    \label{fig:input to model}
\end{figure}

\begin{figure}[ht!]
    \centering
    \includegraphics[scale=0.6]{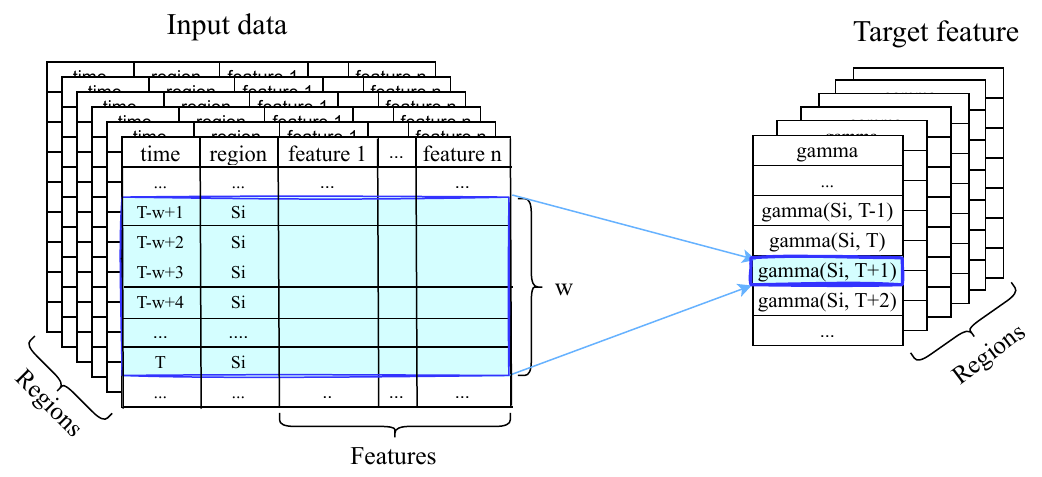}
    \caption{A sequence of \(l(s_i,t),t\in\{T-w+1,T-w+2,\dots,T\}\) as model input and \(Gammaclass_{(s_i,T+1)}\) as its corresponding target feature}
    \label{fig:input to gamma}
\end{figure}

As described in section~\ref{Preprocessing}, we create a vector representation \(l(s,t)\in R^{26}\) for an accident in geographical region \(s\) during time interval \(t\). Each vector has a corresponding \textit{Gammaclass} label that indicates the intensity of delay in traffic flow after an accident. For accident-free data vectors, we label them by $0$, which means no congestion is likely to take place. While this might not be necessarily true in the real-world, it helps to simplify our problem formulation and label data vectors efficiently. 
The model predicts \(gamma\) for time \(t+1\) in region \(s_i\) given sequence of \(w\) previous time intervals in region \(s_i\). Mathematically speaking, \(gamma_{(s_i,T+1)}\) is predicted given a sequence of \(l(s_i,t) ,t\in\{T-w+1,T-w+2,\dots,T\}.\) Figure~\ref{fig:input to model} shows the process of converting the dataset to a 3-D structure of \(l(s_i,t)\) in time and space domain. \(Gammaclass\) is structured in the same order as \(l(s_i,t)\). Figure~\ref{fig:input to gamma} shows an arbitrary input sequence and its corresponding target in our 3-D data structure. 

\begin{figure}
\centering
\begin{subfigure}{.5\textwidth}
  \centering
  \includegraphics[width=0.7\linewidth]{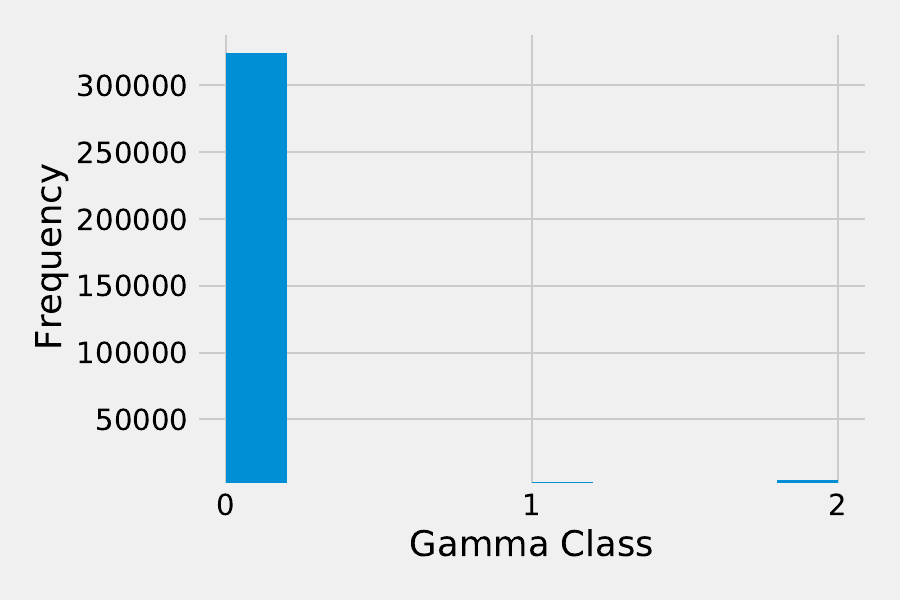}
  \caption{Frequency of gamma classes before RUS}
  \label{fig:sub1}
\end{subfigure}%
\begin{subfigure}{.5\textwidth}
  \centering
  \includegraphics[width=.7\linewidth]{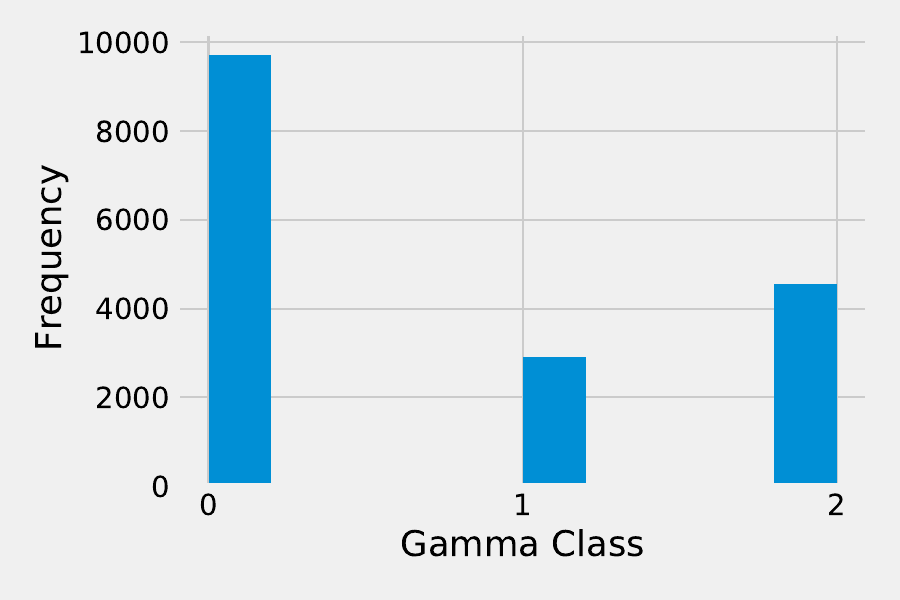}
  \caption{Frequency of gamma classes after RUS}
  \label{fig:sub2}
\end{subfigure}
\caption{Comparing frequency distribution of \(gamma class\) before and after random under sampling}
\label{fig:gammahist}
\end{figure}

We choose accidents from February 2019 to August 2019 as the training time frame. This 27-week time frame includes 13,026 accident representations and 319,194 non-accident representations. Frequency of each \textit{gammaclass} is shown in Figure~\ref{fig:sub1}, indicating that the data is highly imbalanced. Various methods exist to resolve the class imbalance problem in the context of classification or pattern recognition. Examples are (i) removing or merging data in the majority class, (ii) duplicating samples in minority classes, and (iii) adjusting the cost function to make misclassification of minority classes more costly than misclassification of majority instances. In this study, random under-sampling (RUS) method resulted in better outcomes. Hence we use it to mitigate the class imbalance problem of \(gamma  classes\). Using this approach, ratio of accident to non-accident events has increased from $1/43.3$ to $1/1.3$. 
Figure~\ref{fig:sub2} shows the frequency of each \(gamma  class\) after under-sampling. In addition, we assign a weight to each class in the loss function of models (and adjust them during training) to further address the imbalance issue. The following section describes class weighting in more details.

\subsection{Model development}
\begin{figure}[ht!]
    \centering
    \includegraphics[scale=0.68]{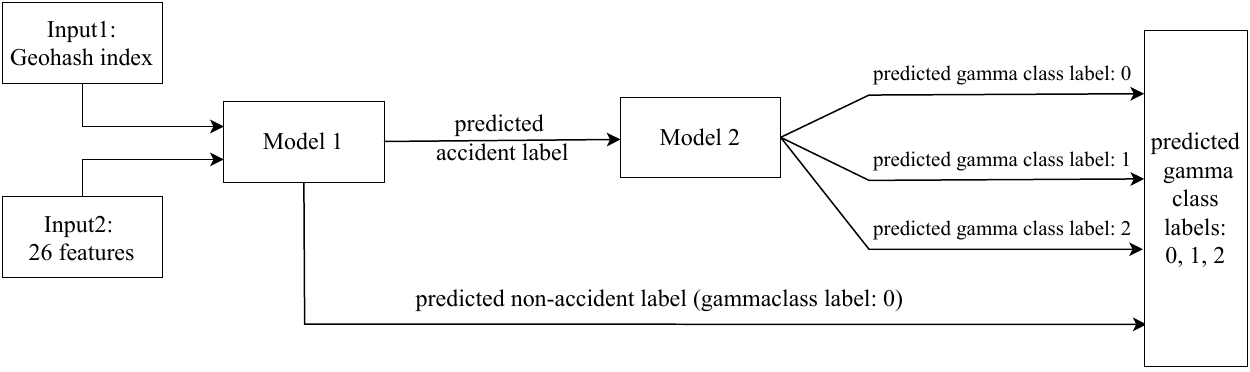}
    \caption{Cascade model overview; the first model predicts whether the next interval would have an accident (i.e., label=1) or not (i.e., label=0). Next, predictions with label=1 are given to the second model that predicts the intensity of accident impact (i.e., \(gammaclass\)). Non-accident predictions of the first model are labeled as \( gammaclass=0\).}
    \label{fig:cascade model}
\end{figure}

Since distinction between accidents and non-accidents is difficult due to complex factors that can affect traffic accident, and some factors that can not be observed and collected in advance (e.g., driver distraction), a model may not perform well on distinction of \(gammaclass\)es using single step prediction (i.e., just by using a single model). Hence, we propose a cascade model that includes two deep neural network components.  It is a cascade model, meaning the output of the first model is served as input for the second model. The first component (or we can call it ``layer'' in our cascade design) focuses on detecting accidents from non-accident events; in other words, the first model detects if there will be an accident in the next two hours given \(w\) previous intervals information. If the first model classifies the input as an accident, then it goes to the second component. The second component focuses on accident impact prediction (i.e., \(gamma classes\)) for accident events, detected by the previous layer. The structure of the proposed accident impact prediction model is shown in Figure~\ref{fig:cascade model}. The structure of the two components in our design is described below. 

\begin{itemize}
    \item [\textbf{A)}]\textbf{Label Prediction}: The First model is a binary LSTM classifier that predicts whether the next interval would have an accident (i.e., label=1) or not (i.e., label=0). In accident prediction it is vital that if an accident is likely to happen, the model can predict it in advance. Therefore, in the first model the focus is on detecting accident events and we use a weighting mechanism for this purpose, such that the weight of the accident class is higher than the weight of non-accident class. Using grid-search, we found optimum weights to be 1 and 3 for non-accident and accident classes, respectively. The structure of the first model is shown in Figure~\ref{fig:lSTM structure}. For this model the input data consists of two components:
    a) index of a given zone \(s_i\) (i.e., \(index_{s_i}\)), and b) \(l(s_i,t)-index_{s_i}\), for $t\in\{T-w+1,T-w+2,\dots,T\}$. Here ``subtraction'' indicates using all features represented by $l(s_i,t)$ except the index of region $s_i$. The first component provides a distributed representation of that cell that encodes essential information in terms of spatial heterogeneity, traffic characteristics, and impact of other environmental stimuli on accident occurrence. It is fed to an embedding layer of size
    \(|R|\times 20\), where R is the set of all grid-cell regions in input dataset. The second component provides other features as a set of \(w\) vectors, each of size 35. This component is fed into two LSTM layers with 12 and 24 neurons, respectively. The output of these layers is then concatenated and fed to two fully connected layers with 25 neurons each. Moreover, batch normalization and dropout layers are added to prevent vanishing gradients and overfitting, respectively.
    
    \begin{figure}[ht!]
        \centering
        \includegraphics[scale=0.46]{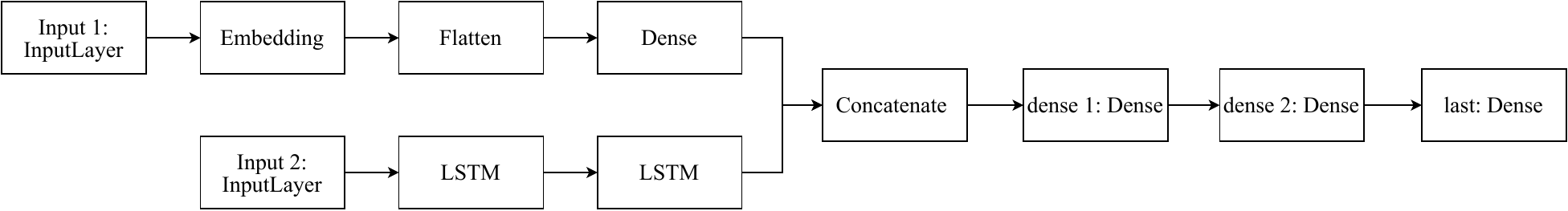}
        \caption{Structure of the first model (layer) in the cascade model. Index of a given zone is fed into an embedding layer to extract latent features for that zone during training. The other features (e.g., weather condition, and congestion data) are fed into LSTM layers. The output of the aforementioned layers is then concatenated and inputted to fully connected layers to predict possibility of an accident. }
        \label{fig:lSTM structure}
    \end{figure}
    
    \item [\textbf{B)}]\textbf{Impact Prediction}: The second model is a 3-class CNN classifier that predicts \(gammaclass\) for next interval. The weight of each class is assigned based on their frequency and importance in our problem setup. Optimum weights are found to be 0.7, 4.5 and 3.5 for \(gammaclass=0\), \(gammaclass=1\) and \(gammaclass=2\), respectively. The structure of the second model is shown in Figure~\ref{fig:CNN structure horiz}. Input for this model is data predicted as an accident event (i.e., label=1) by the first model. The rationale behind considering \(gammaclass=0\) in the second model is that we assume the first model might misclassify some non-accident events, thus this can further ensure the quality of final outcome in case of any misclassifications.  
    
    \begin{figure}[ht!]
        \centering
        \includegraphics[scale=0.30]{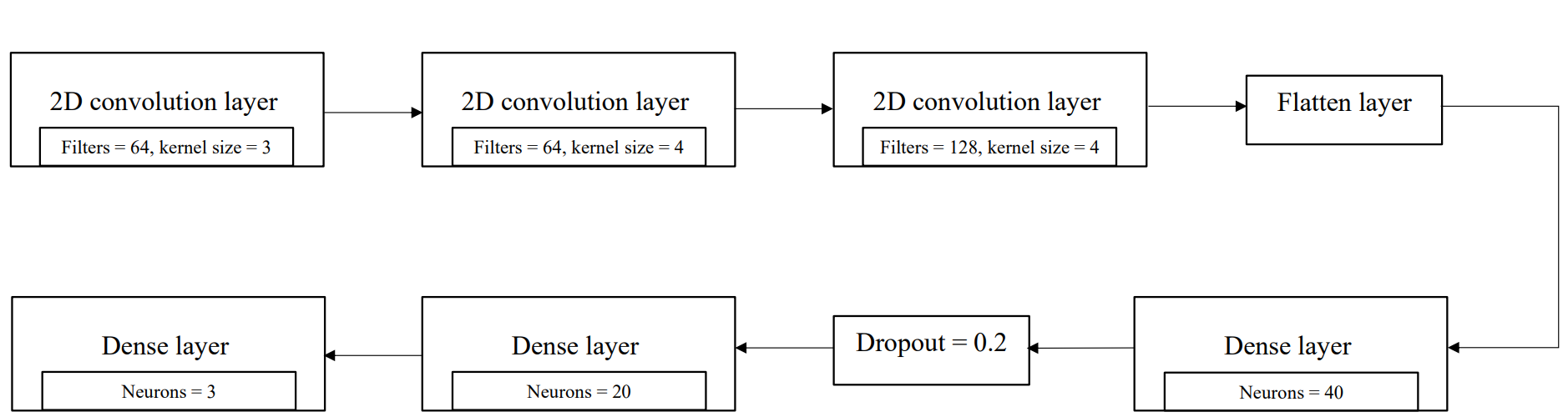}\caption{Structure of the second model (layer) in the cascade model; three consecutive convolutional layers are used to extract spatial latent features from data and then three fully connected layers are used to convert the output of convolutional layers to probability vector for \(gammaclasses\).}
        \label{fig:CNN structure horiz}
    \end{figure}
    
\end{itemize}

To the best of our knowledge, the proposed model is the first cascade model that can be applied in the real-world to classify time intervals as accident and no-accident, and then predict severity of accidents using available data in real-time. Additionally, given the type of the used data, we believe this model can be used in real-world to serve and make compelling predictions. 

%% file: tex/Experiments_and_Results.tex
\section{Experiments and Results}
\label{sec:Experiments and Results}
In this section, we first describe our evaluation metrics, then provide details on experimental setup and lastly present our results followed by discussions. 

\subsection{Evaluation Metrics}\label{Evaluation Metrics}
Failing to predict an accident event (and the consequent impact on traffic flow) is more costly comparing to the case of falsely forecasting traffic delay in an area as a result of a false accident prediction. Hence, we focus on two objectives: 1) maximizing confidence when predicting a non-accident case (i.e., \(gammaclass=0\)), and 2) minimizing possibility of failing to predict accident events (i.e., cases with \(gammaclass=1\) and \(gammaclass=2\)). 
In terms of metrics, these objectives can be translated to 1) ``precision'' for \(gammaclass=0\), and 2) and ``recall'' for \(gammaclass=1\) and \(gammaclass=2\). Precision and recall for multi-class classification are formulated as below:
\[Precision_{class_i} =\frac{M_{ii}}{\sum_i M_{ij}}\]
\[Recall_{class_i} =\frac{M_{ii}}{\sum_j M_{ij}}\]
where \(M_{ij}\) is the number of samples with true class label \(i\) and predicted class label \(j\). Therefore, we focus on precision for class ``0'' and recall for the other two classes for evaluation purpose. That being said, we still need to ensure reasonable recall for class ``0'' and acceptable precision for the other classes. 

\subsection{Experimental Setup}
This section explains the experimental setup and the corresponding test runs. You can find the supplementary materials in our GitHub\footnote{ \url{https://github.com/mahyaqorbani/Accident-Impact-Prediction-using-Deep-Convolutional-and-Recurrent-Neural-Networks}}.
In this study, we used Keras, a Python-based Deep Learning library, to build the prediction models. We choose Adam \cite{kingma2014adam} as the optimizer, given its characteristics to dynamically adjust the learning rate to converge faster and better. To find the optimal models' settings, we performed a grid search over choices of LSTM layers: \{1, 2, 3\}, number of neurons in recurrent layers: \{12,18,24\}, fully-connected layers: \{1, 2\}, and size of fully-connected layers: \{12, 25, 50\} for the first model; and a grid search over choices of convolutional layers: \{1, 2, 3\} and number of filters in each convolutional layer: \{8, 16, 32, 64, 128\} for the second model. The first model (i.e., LSTM) is trained for 150 epochs and the CNN model is trained for 25 epochs. 

\begin{figure}[ht!]
    \centering
    \includegraphics[scale=0.3]{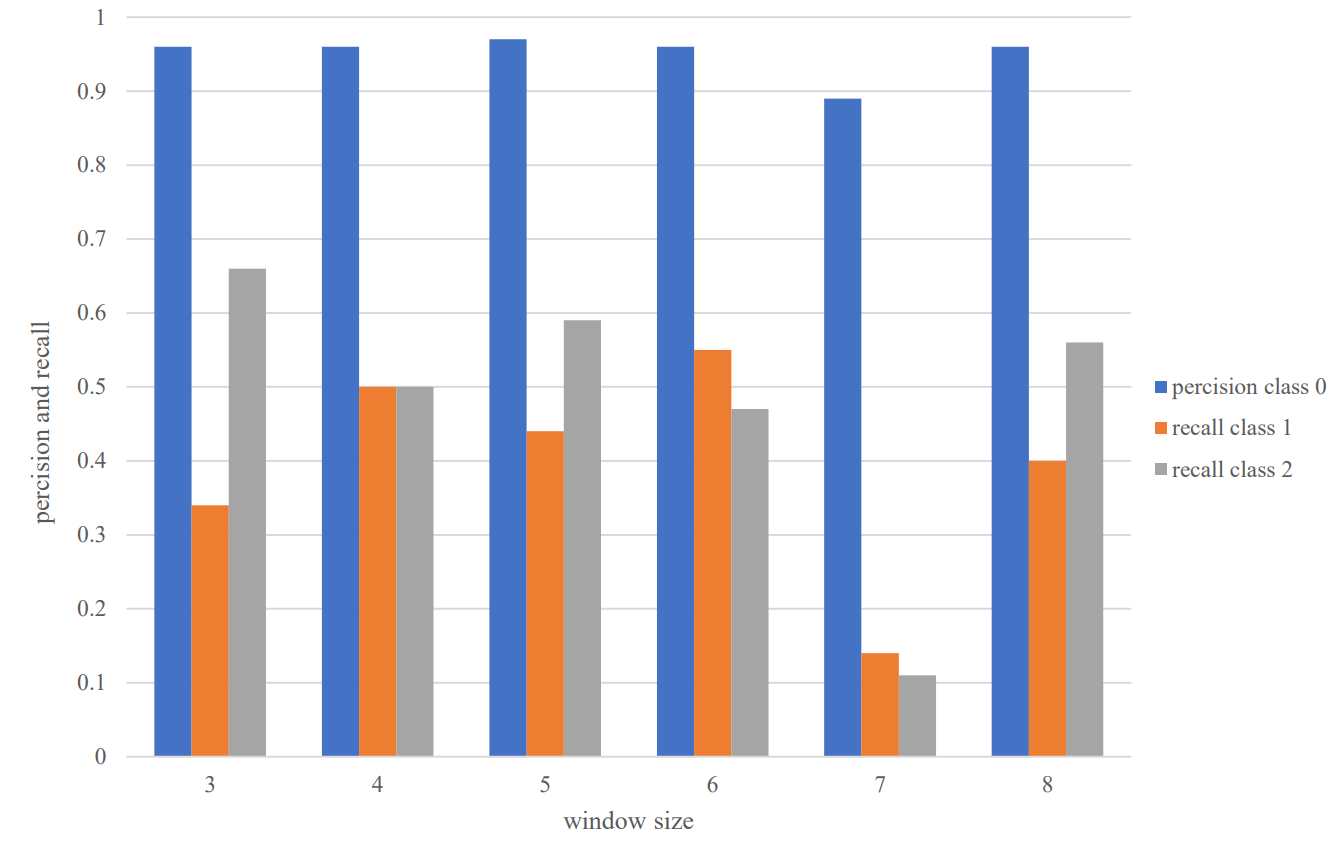}
    \caption{Comparison of evaluation metrics (i.e., \textit{recall} for classes 1 and 2, and \textit{precision} for class 0) based on the cascade model for different choices of input length (or window size)}
    \label{fig:w}
\end{figure}

When building the input vectors, we can use past $w$ intervals to build a vector and pass it to the cascade model to predict a label and a gamma class. But, what is the right choice of $w$? To answer this question, we ran an experiment on the test data to study the metrics introduced in the previous section on different classes for different choices of $w$. The results are shown in Figure~\ref{fig:w}. According to these results, a choice of 4 or 5 for $w$ (which translates to having information from 8 to 10 hours before the accident) seems to be reasonable for this parameter. Here we choose to set $w=4$, since it consistently provides reasonable results over all three gamma classes. 

\subsection{Baseline Models}
In this section, random forest (RF), gradient boosting classifier (GBC), a convolutional neural network (CNN), and a long short-term memory (LSTM) are selected as the baseline algorithms. RF and GBC are traditional models that generally provide satisfactory results on a variety of classification or regression problems. Furthermore, as our proposed cascade model leverages LSTM and CNN components, the use of these two as standalone models seems to be a reasonable choice. We used Scikit-learn and Keras for the off-the-shelf implementation of the baseline models. As for hyperparameters, for the random forest we used 200 estimators, maximum depth=12, minimum sample split=2 and the rest of the parameters were set to defaults. For GBC, we used 300 estimators, learning rate=0.8, maximum depth=2, and the rest of the parameters were set to defaults. The baseline CNN and LSTM models share the same structure as in the proposed cascade model, except for the last layer of the LSTM that uses 3 neurons instead of 2 (since by this model we seek to predict three \(gamma class\)es in a single step). 
\sethlcolor{yellow}
\subsection{Results and Model comparison}
\label{sec:Results}
\begin{figure}[ht!]
    \centering
    \includegraphics[scale=0.28]{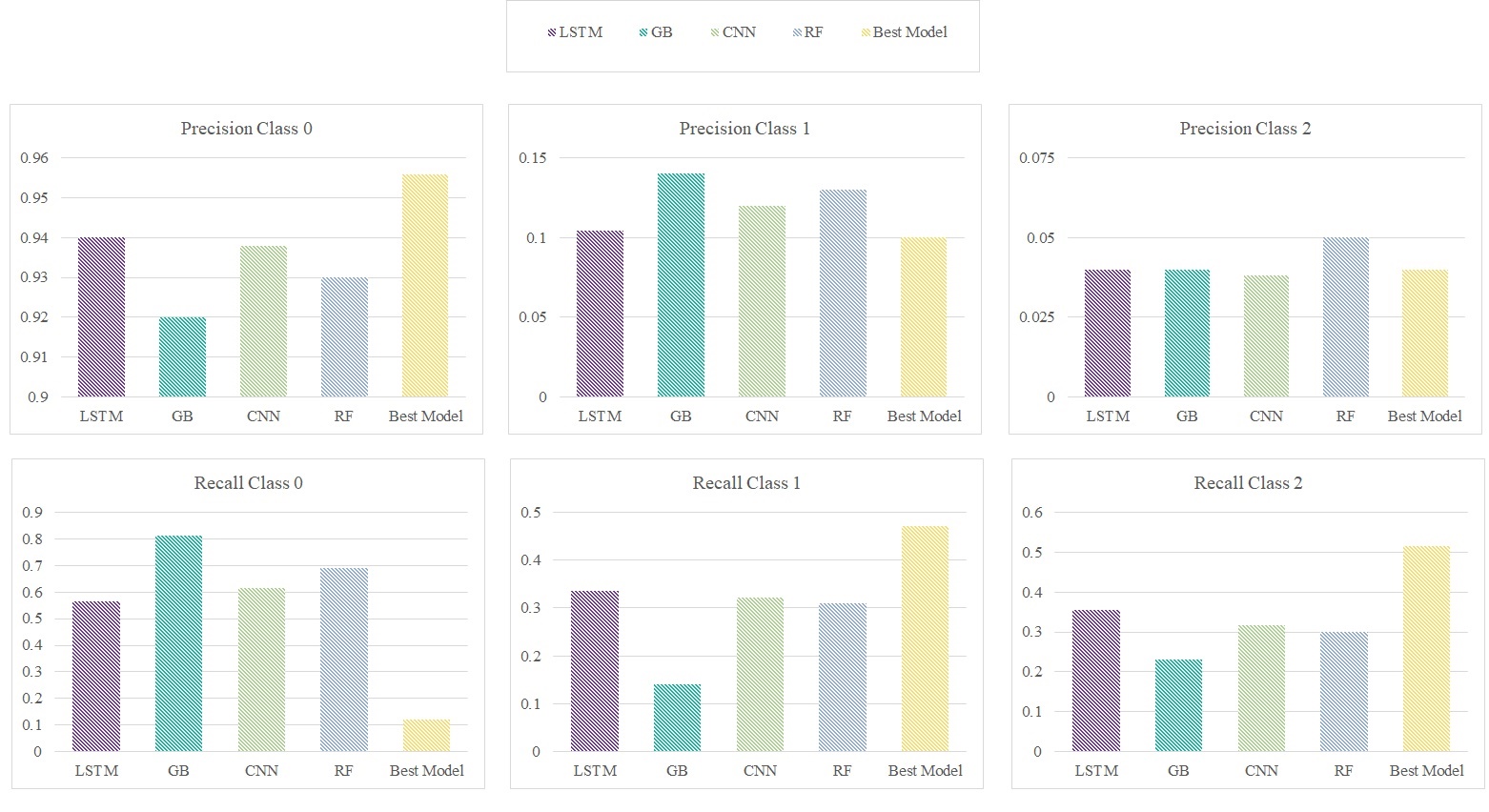}
    \caption{Comparing different models based on precision and recall to predict gamma-classes 0, 1, and 2.}
    \label{fig:metrics versus models.jpg}
\end{figure}

This section provides the evaluation results on test data (from September 2019 to November 2019) to compare our proposal against the two traditional (i.e., RF and GBC) and deep-neural-network models (i.e., CNN and LSTM). A summary of the results is presented in Table~\ref{tab:Results} and Figure~\ref{fig:metrics versus models.jpg}. From these results, we can see that the GBC model performed quite well on detecting non-accident events (i.e., \(gammaclass\)=0) based on precision for this class. However, its results to predict the other two classes are not satisfactory. 

\begin{table}[h!]
 \centering
    \small
    \caption{Accident impact prediction results based on precision for \(gammaclass\)=0 and recall for \(gammaclass\)es 1 and 2}
    \begin{tabular}{|p{0.15\linewidth}|p{0.1\linewidth}|p{0.1\linewidth}|p{0.1\linewidth}|p{0.1\linewidth}|p{0.1\linewidth} |p{0.1\linewidth}|}
        \hline
        \rowcolor{Gray}  \textbf{models} & \textbf{Precision class 0} & \textbf{Precision class 1} & \textbf{Precision class 2} & \textbf{recall class 0} & \textbf{recall class 1} & \textbf{recall class 2} \\
        \hline
        \hline
        Gradient Boosting & 0.92 & 0.14 & 0.04 & 0.81 & 0.14 & 0.23\\
        \hline
        Random Forest & 0.93 & 0.13 & 0.05 & 0.69 & 0.31 & 0.30 \\
        \hline
         LSTM & 0.94 & 0.10 & 0.04 & 0.56 & 0.34 & 0.36\\
       \hline
        CNN & 0.94 & 0.12 & 0.04 & 0.11 & 0.32 & 0.32\\
        \hline
Cascade Model & \textbf{0.96} & 0.10 & 0.04 & 0.31 & \textbf{0.41} & \textbf{0.50}\\
        \hline
    \end{tabular}
    \label{tab:Results}
\end{table}

Further, the RF model does a better job in detecting accidents with medium or high impact. The LSTM model seems to perform better than the other models (except the cascaded model) in detecting accidents with medium or high impact, indicating the importance of taking into account the temporal dependencies to encode input data better. However, CNN's performance is not satisfactory as a single-step model. 

While the proposed cascade model provides satisfactory results in terms of precision for \(gammaclass\)=0, it results in significantly higher recalls to predict the other two classes. These are important observations, because in the case of accident prediction, failing in predicting occurrence of an accident in a zone (i.e., low recall) can result in more serious outcome than falsely predicting accident in a non-accidental zone (i.e., low precision).
Therefore, we generally care more about having higher recall to predict impact for accident events, and also a high precision to decide whether a time-interval is associated with an accident event or not. Please note that while high precision is still necessary for non-zero classes (i.e., cases with reported accident), a real-world framework must ensure acceptable recall for these cases. 

\subsection{Influencing Factors Analysis}
In this section, we conduct a few analysis to study the importance of different factors in our model design (i.e., components and input features). Such analysis can help to improve our input features and model design, all in order to build an effective framework for real-world applications. 
\vspace{2mm}

\begin{figure}
    \begin{subfigure}{.45\textwidth}
      \centering
      \includegraphics[width=0.95\linewidth]{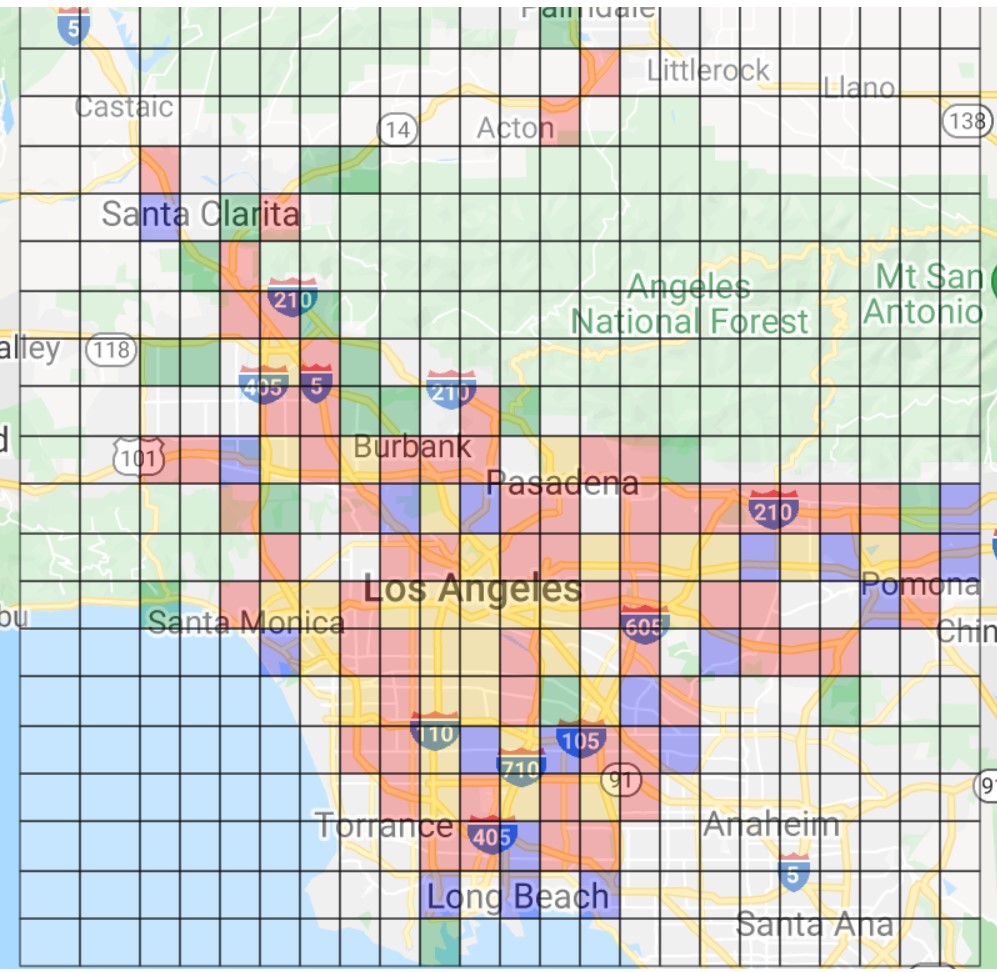}
      \caption{An overview of different clusters 
      of latent representations of regions}
      \label{fig:sub111}
    \end{subfigure}%
    \hfill
    \begin{subfigure}{.5\textwidth}
      \centering
      \includegraphics[width=1.1\linewidth]{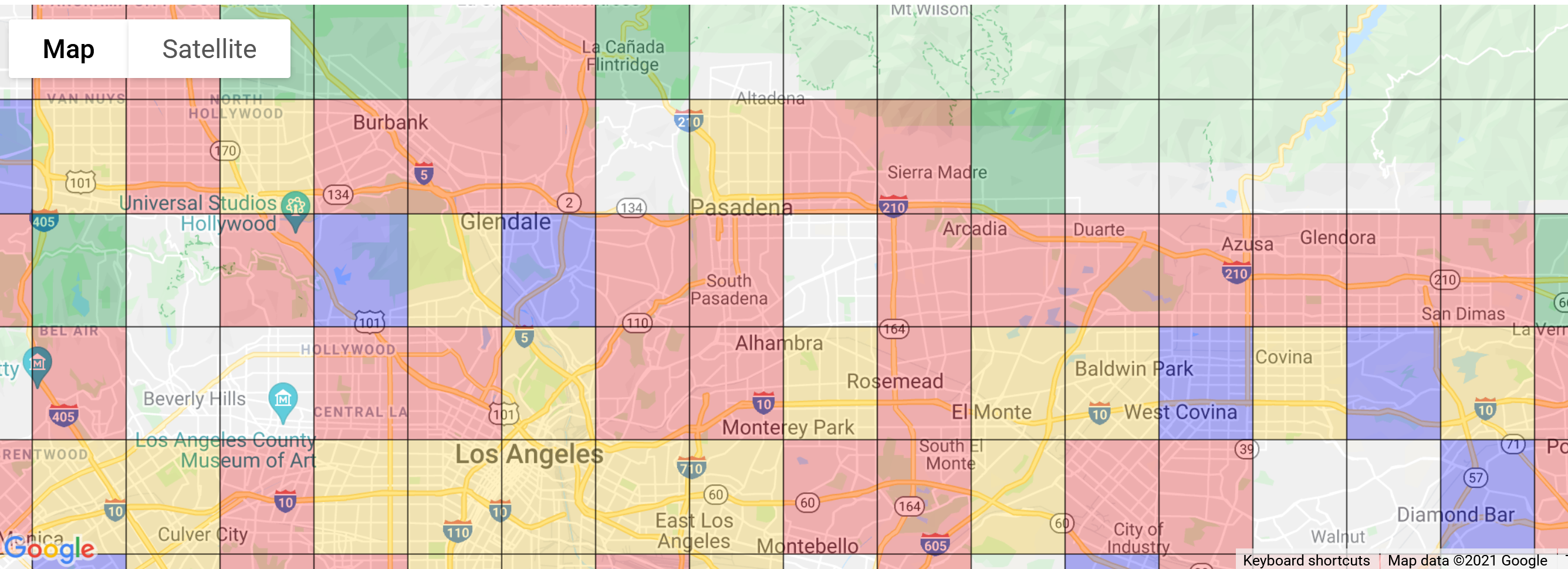}
      \caption{Closer look on Clustering of different regions and their spatial characteristics}
      \label{fig:sub222}
    \end{subfigure}
    \caption{Representation of embedding layer output as different-colored regions on map. Regions represented by transparent color are removed due to lack of enough accident data.}
    \label{fig:geohash clusters}
\end{figure}

\textbf{A) Latent Location Representations:}
As discussed in Sections \ref{undersampling} and \ref{datas}, our dataset covers accident events in an extensive area containing urban road network and high-speed roads.
During the model training, we derive latent representations for each region to encode essential spatiotemporal characteristics. The first analysis is to study quality of derived representations and understand how they can help to distinguish between regions (or zones) with different spatial characteristics.  
We believe that these latent representations can show how well the model has learned spatial factors involved in post-accident impacts. Therefore, we clustered latent (or embedding) vectors into four clusters and specified each cluster with a specific color on a map for better interpretation. Figure~\ref{fig:geohash clusters} shows the results of this analysis. 
Based on Figure~\ref{fig:sub111}, most of the adjacent regions in our study area are found to be in the same cluster, indicating that they share the same accident-related characteristics, which is aligned with principals of urban design. 
As we move away from the center to the outskirts of the county area, we see more heterogeneity in clusters. This can be either a result of less accident reports (i.e., less data for model to train and properly distinguish between regions) or more heterogeneity in road network in suburbs. 
Figure~\ref{fig:sub222} shows more details of different clusters. We can see that urban highways are mainly highlighted as yellow, while areas with sparse road network are highlighted as green. Blue and red regions are mostly urban regions with high density of spatial point of interests (e.g., intersections), and difference between them may be in intensity of traffic congestion caused by accidents. 

In summary, the research findings in this section indicate the ability of our framework to build meaningful latent representations for different regions, that help to better predict accidents and their impact. Moreover, different clusters seem to reasonably represent regions with similar characteristics. The yellow cluster mostly represents regions with a concentrations of arterial highways (for which we can expect higher delays as result of an accident). The green cluster, in contrary, represents regions with marginal and low traffic flow, thus an accident on these regions will not cause as much delay as we should expect for regions in yellow cluster. For the regions in the red cluster, we can expect less delay (or impact) due to lower traffic flow, but higher delay in comparison to the regions in blue cluster.  
Also note that regions with sparse road network and low traffic flow (green regions) are mostly located in suburbs and far from downtown areas.  
\sethlcolor{yellow}

\vspace{2mm}

\begin{figure}[ht!]
    \centering
    \includegraphics[scale=0.33]{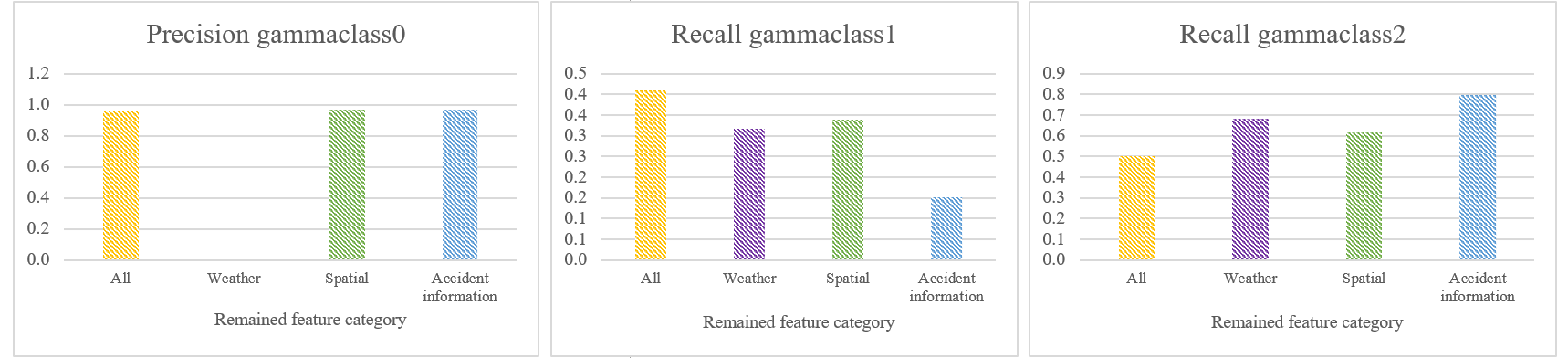}
    \caption{Model performance based on evaluation metrics (Recall of \(gammaclass\)=1, 2 and precision of \(gammaclass\)=0) when using just one feature category (i.e., weather features, spatial features, and accident-related features) in comparison to the case of using all feature categories as input}
    \label{fig:ablation study}
\end{figure}

\textbf{B) Ablation study: }
The second analysis is an ablation study to explore importance of each feature category when used individually for predicting accident impact. Here we study the following categories of features: weather, spatial, and accident information; and compare their results with the case of using all categories of features. Figure~\ref{fig:ablation study} shows the analysis results, and it reveals that removing all but one feature category would significantly degrade the prediction results. 
In comparison to the best model results, using each category alone as an input would result in greater Recall for \(gammaclass\)=2, lower Recall for \(gammaclass\)=1, and almost the same precision for \(gammaclass\)=0 (except for weather category). It is worth noting that using just weather features does not assist in identifying cases where \(gammaclass\)=0. This can either indicate that the quality of weather data in our dataset is not as good as it should be (e.g., it could be finer-grained spatially and temporally), or weather characteristics do not play an important role in the area of our study (throughout the year, we do not usually see significant shifts in weather condition in Los Angeles area\footnote{See \url{https://weatherspark.com/y/1705/Average-Weather-in-Los-Angeles-California-United-States-Year-Round} for more details.}).

%% file: tex/conclusion.tex
\section{Conclusion and Future Work}
\label{sec:conclusion}
Traffic accidents are serious public safety concerns, and many studies have focused on analyzing and predicting these infrequent events. However, existing studies suffer from employing
extensive data that may not be easily available to other researchers, or in real-time to be utilized for real-world applications. Additionally, they fail to establish an appropriate criterion for determining the impact of accidents. To overcome these limitations, this study proposes a cascade model that combines LSTM and CNN components for real-time traffic accident prediction. The model detects future accidents and assesses their impact on surrounding traffic flow using a novel metric called gamma. Four complementary datasets, including accident data, congestion data, weather data, and spatial data, are utilized to construct the input data. Through extensive experiments conducted using data from Los Angeles county, we demonstrate that our proposed model outperforms existing approaches in terms of precision for cases with minimal impact (i.e., no reported accidents, \(gammaclass\)=0) and recall for cases with significant impact (i.e., reported accidents, \(gammaclass\)=1 and 2).

This study has several implications for future research. Firstly, we suggest expanding the framework to incorporate data from neighboring regions to predict accident impact for a target region, mitigating data sparsity. Additionally, the inclusion of satellite imagery data to enhance the model's ability to distinguish between regions and capture spatial information could also improve performance.
\sethlcolor{green}

While our work makes important contributions, it is important to note its limitations. Our model currently only classifies accident impacts into three broad categories. Future research could benefit from more granular classifications or assigning probabilities to each accident's potential impact on road conditions. Additionally, the accuracy of our results may have been limited by the accessibility and quality of the weather data we used from a private source, which was gathered from only four airports. To enhance the accuracy of our model, future research could incorporate radar data or other more comprehensive weather data sources. Addressing these limitations will be essential to advancing our understanding of the factors that contribute to road accidents and improving road safety.

Overall, this study provides a promising approach to real-time traffic accident prediction using a novel metric for measuring accident impact. The proposed framework has the potential to be applied in real-world settings to enhance public safety and traffic management.

%% file: tex/Interest.tex
\section{Declaration of Competing Interest}
The authors declare that they have no known competing financial interests or personal relationships that could have appeared to influence the work reported in this paper.

%% file: tex/Data_Availability.tex
\section{Data Availability}
\sethlcolor{yellow}
All data generated or analysed during this study are included in this published article, ``Short and Long-term Pattern Discovery Over Large-Scale Geo-Spatiotemporal Data'' \cite{moosavi2019short}. The Weather Condition Dataset and POI Dataset are not available publicly, but you can refer to the paper mentioned above to get insightful information about how to collect such a dataset.